\documentclass[sigconf]{acmart}

\usepackage{graphicx}
\usepackage{amsmath}

\usepackage{amssymb}
\usepackage{booktabs}
\usepackage{multirow}
\usepackage{enumitem}
\usepackage{xcolor}
\usepackage{array}
\newcommand{\PreserveBackslash}[1]{\let\temp=\\#1\let\\=\temp}
\newcolumntype{C}[1]{>{\PreserveBackslash\centering}p{#1}}
\newcolumntype{R}[1]{>{\PreserveBackslash\raggedleft}p{#1}}
\newcolumntype{L}[1]{>{\PreserveBackslash\raggedright}p{#1}}
	
	\AtBeginDocument{%
		}

\begin{document}
		
		\title{Open-Vocabulary Object Detection via Scene Graph Discovery}
		
		
\author{Hengcan Shi}
\affiliation{%
	  \institution{Department of Data Science \& AI, Monash University}
	  \country{Australia}}
\email{hengcan.shi@monash.edu}

\author{Munawar Hayat}
\affiliation{%
	\institution{Department of Data Science \& AI, Monash University}
	\country{Australia}}
\email{munawar.hayat@monash.edu}

\author{Jianfei Cai}
\affiliation{%
	\institution{Department of Data Science \& AI, Monash University}
	\country{Australia}}
\email{jianfei.cai@monash.edu}

		
		\begin{abstract}
			In recent years, open-vocabulary (OV) object detection has attracted increasing research attention. Unlike traditional detection, which only recognizes fixed-category objects, OV detection aims to detect objects in an open category set. Previous works often leverage vision-language (VL) training data (e.g., referring grounding data) to recognize OV objects. However, they only use pairs of nouns and individual objects in VL data, while these data usually contain much more information, such as scene graphs, which are also crucial for OV detection. In this paper, we propose a novel Scene-Graph-Based Discovery Network (SGDN) that exploits scene graph cues for OV detection. Firstly, a scene-graph-based decoder (SGDecoder) including sparse scene-graph-guided attention (SSGA) is presented. It captures scene graphs and leverages them to discover OV objects. Secondly, we propose scene-graph-based prediction (SGPred), where we build a scene-graph-based offset regression (SGOR) mechanism to enable mutual enhancement between scene graph extraction and object localization. Thirdly, we design a cross-modal learning mechanism in SGPred. It takes scene graphs as bridges to improve the consistency between cross-modal embeddings for OV object classification. Experiments on COCO and LVIS demonstrate the effectiveness of our approach. Moreover, we show the ability of our model for OV scene graph detection, while previous OV scene graph generation methods cannot tackle this task.
		\end{abstract}
		
		%
		
		
		
		\maketitle

		\begin{figure}
			\centerline{\includegraphics[scale=0.45]{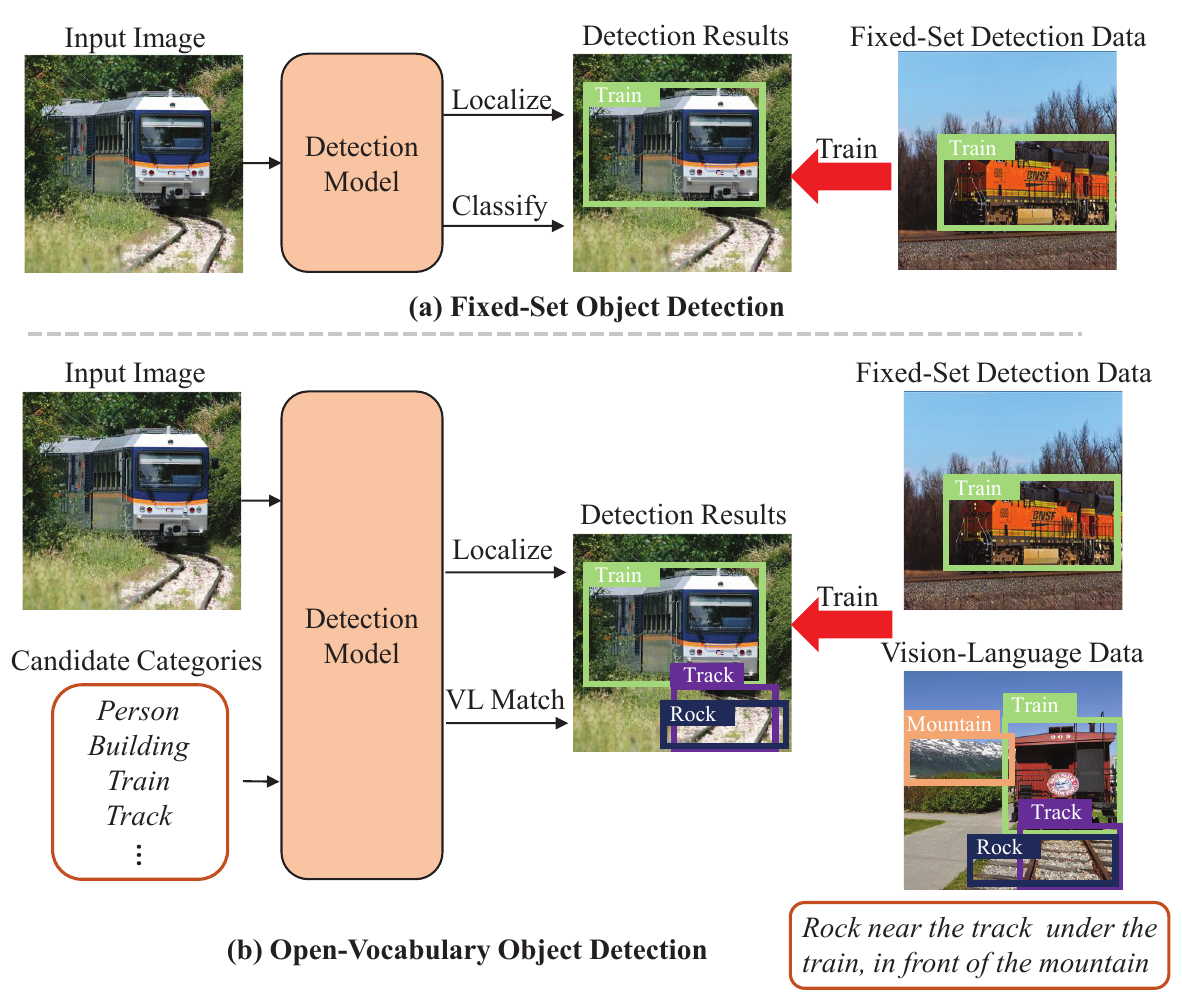}}
			\caption{(a) Traditional object detection methods only classify and localize objects in a fixed category set. (b) OV object detection expects to recognize various object categories, not only objects seen in detection training data but also unseen objects. To this end, during training, OV methods often add VL training data or distill knowledge from pre-trained VL models to recognize OV objects. During inference, they first localize objects and then classify them by VL matching with a set of candidate categories in the target application.}
			\label{fig_introduction}
		\end{figure}
		
		\section{Introduction}
		Object detection is an important and fundamental problem in computer vision, which serves as a crucial step for many higher-level tasks, such as scene understanding \cite{ma2021implicit, shi2018boosting}, image captioning \cite{tang2021clip4caption, chen2021towards} and cross-modal retrieval \cite{wang2022point, cao2022correspondence, qiu2020language, li2021bottom}. Traditional object detection expects to classify and localize objects in a fixed category set, as shown in Fig.~\ref{fig_introduction}(a). Consequently, users have to continually retrain the model to fit different real-world applications, because different applications normally involve varying category sets. Hence, open-vocabulary (OV) object detection \cite{du2022learning, gu2021open, ma2022open} has attracted increasing attention in recent years, where the model is trained to recognize an open set of object categories and thus can be directly used for diverse applications.
		
		However, object detection training data only contain objects of limited categories, and thus the key challenges in OV detection are how to discover, classify and localize unseen objects. For object discovery, unseen objects may be treated as `\emph{background}' by detection networks, and thus no proposal is generated for them. Similarly, without corresponding training data, detection networks are also hard to accurately localize and classify unseen objects.
		OV detection methods usually tackle these problems by introducing vision-language (VL) information, as illustrated in Fig.~\ref{fig_introduction}(b), because language involves various objects. 
		The existing works use three types of methods to incorporate VL information.
		The first is pre-training-based methods \cite{zareian2021open, gu2021open, ma2022open, du2022learning, zang2022open}, which distill knowledge (e.g., feature spaces) from pre-trained VL models to discover and classify OV objects, and leverage fixed-set detection data to train modules for object localization. Nevertheless, pre-trained models reduce the flexibility of these methods. They have to encode their features into the pre-trained feature space and cannot flexibly adjust them.
		The second type is weakly-supervised methods \cite{zhou2022detecting, lin2022learning, huynh2022open, zhao2022exploiting}, which first generate pseudo OV detection labels from image-level VL data, and then train detection networks with these pseudo labels. Nonetheless, these methods suffer from the problem of inevitable noises in pseudo labels.
		The third type \cite{kuo2022findit,li2022grounded} reformulates object detection as referring grounding problem, and thus can leverage referring grounding training data to simultaneously enable OV object discovery, classification and localization. Such approaches avoid the noise in weakly-supervised methods and are more flexible than pre-training-based methods. Despite significant progress made by these methods, they only extract object names in referring expressions, but ignore other rich language information. 
		
		As shown in Fig.~\ref{fig_introduction}(b), language expressions usually contain not only object names but a mass of object relations (e.g., `\emph{near}' and `\emph{under}'), which are also important cues for OV object detection. Firstly, unseen objects can be better discovered by relation cues. For example, when a network finds an `\emph{under train}' relation in the image, there might be an object under the train, even if the network has not seen this object before.
		Secondly, relations can also improve object classification accuracy. For instance, the object under trains is probably `\emph{track}'.
		Thirdly, as many relations describe object positions, such as `\emph{under}' and `\emph{in front of }', they are helpful for object localization.
		
		Based on these observations, we propose a novel Scene-Graph-Based Discovery Network (SGDN) to exploit object relations for OV object detection. Specifically, we first present a scene-graph-based transformer decoder (SGDecoder) to model both objects and their relations, i.e., scene graphs. In SGDecoder, a sparse scene-graph-guided attention (SSGA) module is designed to embed scene graph cues into object representations for OV object discovery, localization and classification. Based on these representations, scene-graph-based prediction (SGPred) is proposed to predict OV detection results, including bounding boxes and categories. For bounding box prediction, we build a scene-graph-based offset regression (SGOR) mechanism, where object localization and scene graph modeling are mutually boosted. For classification, we present a cross-modal learning method that leverages scene graphs to improve the consistency between cross-modal object embeddings. SGPred also generates relation predictions to better learn relation information. 
		
		Our major contributions can be summarized as follows. 
		\begin{enumerate}
			\item We propose a novel scene-graph-based OV object detection network, SGDN. To the best of our knowledge, this is the first work that exploits scene graph cues for OV object detection.
			\item We present SGDecoder including an SSGA module to model scene graphs for OV object discovery, localization and classification. An SGPred method with SGOR and cross-modal learning mechanisms are designed to improve OV predictions based on scene graph cues.
			\item Our SGDN outperforms many previous state-of-the-art methods on two common OV detection datasets, COCO and LVIS. Meanwhile, SGDN can also generate OV scene graph detection results, while previous OV scene graph generation methods cannot.
		\end{enumerate}
		
		\section{Related Work}
		
		\textbf{Open-Vocabulary Object Detection.} The existing OV detection works can be generally categorized into three types. The first is pre-training-based OV detection. They are trained with fixed-set detection data to localize objects while incorporating VL pre-training models to recognize OV objects. OVR-CNN \cite{zareian2021open} uses image caption data to train a model as the pre-training. Many other methods \cite{gu2021open, ma2022open, du2022learning} employ off-the-shelf pre-trained models, such as CLIP \cite{radford2021learning}. 
		F-VLM \cite{kuo2022f} adds classification and localization heads to CLIP, and fine-tunes the model on detection data. ViLD \cite{gu2021open} and OV-DETR \cite{zang2022open} distill knowledge from CLIP to detection networks to generate OV detection results. 
		HierKD \cite{ma2022open} extracts multi-scale features for better OV detection. DetPro \cite{du2022learning} incorporates prompt learning to boost performance, and Promptdet \cite{feng2022promptdet} further enhances the prompt learning to region level.
		These methods successfully leverage VL pre-training to improve their OV recognition ability. Nevertheless, the flexibility of these methods is limited by pre-training models, because they have to encode their features into the pre-trained features space and cannot flexibly adjust them. 
		
		The second type is weakly-supervised approaches. They leverage large-scale image-level supervisions (such as image classification and image caption data) to train detection models, to address the issue of lacking OV dense annotations.
		To train an OV detection model, Detic \cite{zhou2022detecting} extracts pseudo bounding boxes from image classification datasets with up to 21K categories. Gao \emph{et al.} \cite{gao2021towards}, RegionCLIP \cite{zhong2022regionclip}, and VL-PLM \cite{zhao2022exploiting} generate pseudo bounding boxes from CLIP by class activation map (CAM) or pre-trained RPN \cite{ren2015faster}. Rasheed \emph{et al.} \cite{rasheed2022bridging} first generate object detection results and then restore image-level results from these object-level predictions. In this way, they can use image-level supervision to train the model. 
		VLDet \cite{lin2022learning} uses image-caption supervision by aligning each noun in the caption to object proposals, where object proposals are extracted by pre-trained detectors. However, there are inevitable noises during weakly-supervised training, which limits the performance.
		
		\begin{figure*}
			\centerline{\includegraphics[scale=0.52]{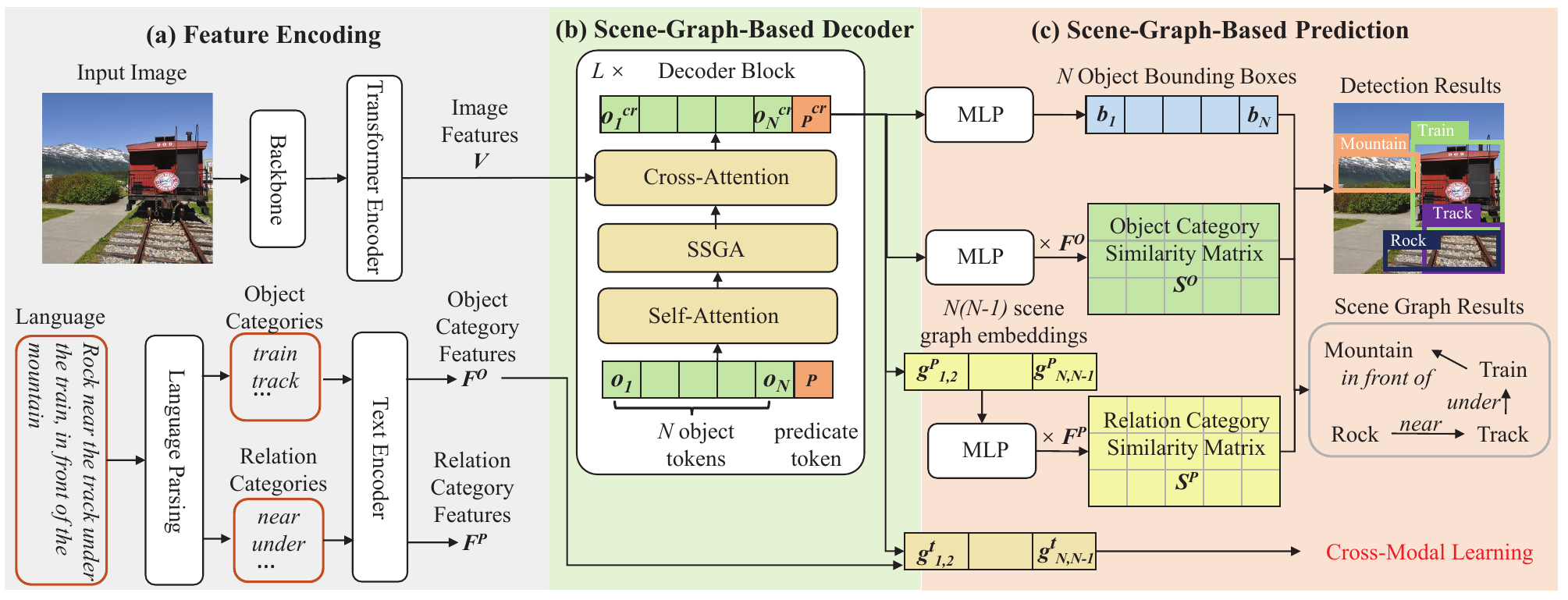}}
			\caption{Illustration of our SGDN. Our model consists of three parts. (a) Feature encoding, including an image backbone encoder, an image transformer decoder and a text encoder, extracts image and text features. Language parsing tools are used during training to extract object and relation categories. (b) Scene-graph-based decoder (SGDecoder), containing our SSGA module, self- and cross-attention, models object and relation embeddings. (c) Scene-graph-based prediction (SGPred) to predict object bounding boxes, object and relation categories. We also present cross-modal learning and SGOR (see Fig.~\ref{fig_method2}) for better category and bounding box prediction.}
			\label{fig_method}
		\end{figure*}
		
		The third type, grounding-based works, points out the high similarity between OV detection and referring grounding, and uses grounding frameworks to tackle OV detection. Since grounding training data includes bounding box annotations for diverse objects, FindIt \cite{kuo2022findit} combines referring comprehension and object detection data to train a grounding model, which shows good OV detection ability. X-DETR \cite{cai2022x} reformulates detection and grounding as an instance-text alignment problem, and designs an alignment network for both tasks. GLIP \cite{li2022grounded} enhances the VL interaction in the alignment framework, and also extracts millions of pseudo grounding labels from image caption data to boost the training. GLIPv2 \cite{zhang2022glipv2} extends GLIP for more tasks, such as image captioning and visual question answering. Nevertheless, they ignore object relation information in referring expressions, which are also important for OV detection. Unlike them, we exploit object relations to improve OV object discovery, classification and localization.
		
		\textbf{Object Detection and Scene Graph.} Early scene graph generation methods \cite{lu2016visual, xu2017scene} employ pre-trained object detectors to extract bounding boxes for relation prediction. They do not optimize object detectors. Recent works \cite{li2022sgtr, shit2022relationformer, li2021bipartite, li2017scene, shi2021simple, li2022integrating} simultaneously optimize object detection and relation predictions. These scene graph approaches are foundations of our work. However, they more focus on relation prediction from object detection cues, rather than exploring relation cues for object detection. Several works \cite{liu2018structure, yang2020graph, lyu2020vtgraphnet} leverage scene graph cues for object detection and referring grounding. SIN \cite{liu2018structure} implicitly models object relations without any relation supervision for fixed-set object detection. SGMN \cite{yang2020graph} and vtGraphNet \cite{lyu2020vtgraphnet} disassemble complex referring expressions into scene graphs, and thus simplify the reasoning for referring grounding. Different from them, we exploit scene graphs for OV object detection. We propose modules that leverage scene graph cues to discover, classify and localize OV objects.
		
		Some works \cite{zhong2021learning, he2022towards} study the OV scene graph generation problem. Zhong \emph{et al.} \cite{zhong2021learning} design a weakly-supervised method and leverage image caption data to capture OV knowledge. 
		SVRP \cite{he2022towards} uses VL data to pre-train a model for OV relation recognition, and then designs a prompt to predict relations between two objects. 
		However, these methods also focus on relation prediction and have no mechanism for OV object detection. As a result, they can only tackle the OV predicate classification and scene graph classification tasks, while cannot generate OV scene graph detection results. Compared with them, our model aims at OV detection with relation cues, and is able to deal with the OV scene graph detection problem.
		
		\section{Proposed Method}
		\subsection{Problem Definition and Method Overview}
		The inputs of OV object detection are an image and a number of text, as shown in Fig.~\ref{fig_introduction}(b). During inference, the text are usually $C$ candidate object categories. OV detection networks output object proposals (bounding boxes) from the image, and determine the category of each object by matching the object embedding with $C$ candidate category embeddings. During training, the text can be any language description corresponding to objects in the image. By learning with these VL data, OV detection networks are able to recognize various objects.
		
		Previous works \cite{lin2022learning, li2022grounded} only use nouns (i.e., individual objects) from language descriptions, while ignoring other useful information such as object relations. Objects and relations can compose scene graphs, which provide important cues for OV object discovery, classification and localization. A scene graph triplet is formed as `subject-predicate(relation)-object', where `subject' and `object' are two objects and `predicate' is the relation between them. In this paper, we propose an SGDN that exploits scene graphs for OV object detection. 
		
		Our SGDN consists of three components as illustrated in Fig. \ref{fig_method}. 
		The first is Feature Encoding that extracts the embeddings of the input image and text. 
		The second is SGDecoder to generate embeddings of objects and relations. In SGDecoder, we propose an SSGA module to enrich object embeddings by scene graph information to improve OV object discovery, classification and localization. 
		The final component is SGPred that generates object bounding boxes and categories, as well as relation categories. In SGPred, an SGOR (Fig.~\ref{fig_method2}) is used to mutually refine scene graph extraction and bounding boxes prediction. We also propose cross-modal learning, which takes scene graphs as bridges to enhance the consistency between cross-modal embeddings.
		Next, we introduce each module in detail.
		
		\subsection{Feature Encoding} \label{Encoding}
		\textbf{Image encoder.} We leverage the common transformer-based architecture \cite{zhu2021deformable} for object detection, where the image encoder includes a backbone encoder (e.g., ResNet \cite{he2016deep}) and a transformer encoder \cite{zhu2021deformable}. The output of our encoder is an image feature map $ \mathbf{V} \in \mathbb{R}^{N_{v} \times D_{v}}$, where $N_{v}$ is the number of image patches, and $D_{v}$ is the dimension. 
		
		\textbf{Text encoder.}  We extract textual embeddings by a pre-trained text encoder (e.g., BERT \cite{kenton2019bert} or RoBERTa \cite{liu2019roberta}). Since we generate both object and relation predictions, our input text contains two parts: object categories and relation categories. 
		
		During inference, there are $C+1$ object categories, including $C$ candidate object categories in the target application and an additional `\emph{no object}' category to recognize false proposals. The text encoder generates a feature map $\mathbf{F}^{o} \in \mathbb{R}^{(C+1) \times D}$ for object categories, in which each feature vector encodes an object category, and $D$ is the feature dimension.
		Similarly, we have $M+1$ relation categories, i.e., $M$ candidate relation categories in the target application and an additional `\emph{no relation}' category.  
		$\mathbf{F}^{p} \in \mathbb{R}^{(M+1) \times D}$ represents the output relation category feature map.
		Note that if an object or relation category contains multiple words, our text encoder can generate one feature vector of the entire category. 
		
		During training, we first use language parsing tools \cite{schuster2015generating} to extract nouns and relations of nouns from the language expression. Then, we take all nouns in this expression as the candidate object categories, and also add the `\emph{no object}' category. All relations in this expression as well as the `\emph{no relation}' category are treated as our relation categories. 
		
		Our method leverages SGDecoder to model scene graphs, and candidate relation categories are only used for relation prediction. If training data (e.g., fixed-set detection data) or target applications only require object detection results, our model can avoid these relation inputs and skip the relation prediction and cross-modal learning parts in SGPred.
		
		\subsection{Scene-Graph-Based Decoder} \label{Decoder}
		After feature encoding, we build an SGDecoder to extract object and predicate embeddings. The inputs of our decoder are $N$ object tokens $\{\mathbf{o}_{n} \in \mathbb{R}^{D}\}_{n=1,...N}$ and a predicate token $\mathbf{p} \in \mathbb{R}^{D}$, where $D$ is the dimension of each token. Each object token $\mathbf{o}_{i}$ represents an object in the image. As investigated in previous scene graph generation works \cite{shit2022relationformer}, the relation and scene graph between the $i$-th and $j$-th objects can be represented by the concatenation of the `subject' embedding $\mathbf{o}_{i}$, the predicate embedding $\mathbf{p}$ as well as the 'object' embedding $\mathbf{o}_{j}$; and the predicate embedding $\mathbf{p}$ can be shared for all object pairs. Therefore, we only use one relation token to capture object relations. Our decoder generates object and predicate embeddings from these $N+1$ object and relation tokens and the image feature $\mathbf{V}$.
		
		\textbf{Self-attention.} The decoder contains $L$ blocks and each block includes a self-attention, an SSGA module and a cross-attention. The self-attention models long-range dependencies among object and predicate tokens, and updates these tokens as follows:
		\begin{equation}
			\begin{split}
				\{\mathbf{o}^{sl}_{1}, ..., \mathbf{o}^{sl}_{N}, \mathbf{p}^{sl}\} = Attn(&query: \{\mathbf{o}_{1}, ..., \mathbf{o}_{N}, \mathbf{p}\}, \\
				& key: \{\mathbf{o}_{1}, ..., \mathbf{o}_{N}, \mathbf{p}\}, \\
				& value: \{\mathbf{o}_{1}, ..., \mathbf{o}_{N}, \mathbf{p}\})
			\end{split}
		\end{equation}
		where $Attn(\cdot)$ is a multi-head attention model \cite{vaswani2017attention}. We take our object and predicate tokens $\{\mathbf{o}_{1}, ..., \mathbf{o}_{N}, \mathbf{p}\}$ as queries, keys and values in this attention model. $\{\mathbf{o}^{sl}_{1}, ..., \mathbf{o}^{sl}_{N}, \mathbf{p}^{sl}\}$ are the updated object and predicate tokens, where long-range dependencies are embedded, and every token is also $D$-dimension.
		
\begin{figure}
			\centerline{\includegraphics[scale=0.5]{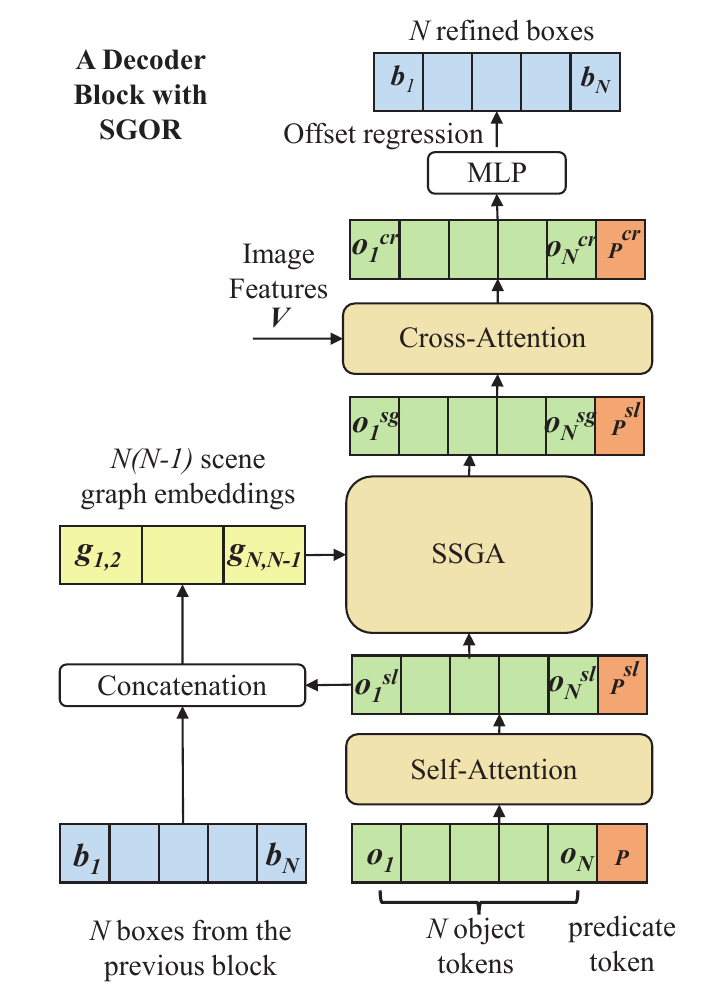}}
			\caption{Illustration of a decoder block and the SGOR mechanism. In the first block, the inputs are initialized bounding boxes, object and predicate tokens. The block first uses self-attention to model dependencies among object and predicate tokens. Scene graph embeddings are then generated based on these tokens and bounding boxes. An SSGA module embeds scene graph information into object embeddings for better OV object detection. Cross-attention further integrates image information into these embeddings. The SGOR mechanism leverages MLPs to predict bounding box offsets and refine bounding boxes. The refined bounding boxes, object and predicate tokens are inputs of the next decoder block. In this way, scene graph extraction and object localization can be mutually improved.}
			\label{fig_method2}
		\end{figure}

		\textbf{The SSGA module.} We propose an SSGA module that further embeds scene graph information into object tokens to improve OV object discovery, classification and localization. Specifically, as shown in Fig.~\ref{fig_method2}, we first generate scene graph embeddings:
		\begin{equation} \label{eqn_sg}
			\mathbf{g}_{i,j} = [\mathbf{o}^{sl}_{i}, \mathbf{b}_{i}, \mathbf{p}^{sl}, \mathbf{o}^{sl}_{j}, \mathbf{b}_{j}],  \quad  i,j = 1, ..., N \ and \ i \neq j
		\end{equation} 
		where $[\cdot]$ means token concatenation. $\mathbf{b}_{i}, \mathbf{b}_{j} \in \mathbb{R}^{4}$ are bounding boxes of the $i$-th and $j$-th objects, respectively. We integrate bounding box information to generate more powerful scene graph embeddings. The details of bounding boxes will be introduced in Sec.~\ref{Prediction}.
		The scene graph embedding $\mathbf{g}_{i,j} \in \mathbb{R}^{3D+8}$ encodes the information of the $i$-th, $j$-th objects as well as their relation. All scene graph embeddings can compose a scene graph matrix $\mathbf{G} \in \mathbb{R}^{N(N-1) \times (3D+8)}$. 
		
		Our SSGA module then leverages an attention model to embed scene graphs into object tokens as
		\begin{equation}
			\begin{split}
				\{\mathbf{o}^{sg}_{1}, ..., \mathbf{o}^{sg}_{N}\} = SAttn(&query: \{\mathbf{o}^{sl}_{1}, ..., \mathbf{o}^{sl}_{N}\}, \\
				& key: \mathbf{G}, \\
				& value: \mathbf{G})
			\end{split}
		\end{equation}
		where we take object tokens as queries, and treat scene graph embeddings as keys and values. $SAttn(\cdot)$ means a sparse attention model. To reduce computational costs, we only calculate attention between each object token and its related scene graph embeddings, i.e., this object acts as the `subject' or `object' in the scene graph embedding. For example, for the $n$-th object token, we only compute attention between $\mathbf{o}^{sl}_{n}$ and $\{\mathbf{g}_{n,j}, \mathbf{g}_{j,n}\}_{j=1,...,N \ and \ j \neq n}$. The attention model incorporates scene graph guidance into object tokens and updates object tokes into $\{\mathbf{o}^{sg}_{n} \in \mathbb{R}^{D} \}_{n=1,...N}$. In this way, although our model has not seen some objects in training data, it can discover them from scene graph cues. Moreover, these scene graph cues are also helpful for object classification and localization.
		
		\textbf{Cross-attention.} The cross-attention takes object and predicate tokens as queries, while using the image feature map $\mathbf{V}$ as keys and values to integrate visual information into these tokens:
		\begin{equation}
			\begin{split}
				\{\mathbf{o}^{cr}_{1}, ..., \mathbf{o}^{cr}_{N}, \mathbf{p}^{cr}\} = Attn(&query: \{\mathbf{o}^{sg}_{1}, ..., \mathbf{o}^{sg}_{N}, \mathbf{p}^{sl}\}, \\
				& key: \mathbf{V}, \\
				& value: \mathbf{V}).
			\end{split}
		\end{equation}
		The output embeddings $\{\mathbf{o}^{cr}_{1}, ..., \mathbf{o}^{cr}_{N}, \mathbf{p}^{cr}\}$ of each decoder block embed image, object and scene graph information. All embeddings are $D$-dimension vectors and can be further refined by the next decoder block.
		
		\subsection{Scene-Graph-Based Prediction} \label{Prediction}
		Based on our object and predicate embeddings $\{\mathbf{o}^{cr}_{1}, ..., \mathbf{o}^{cr}_{N}, \mathbf{p}^{cr}\}$, three types of results can be predicted, i.e., object bounding boxes and categories, as well as relation categories).
		
		\textbf{Object bounding box prediction.} Let $\mathbf{b}_{n} = [x_{n}, y_{n}, w_{n}, h_{n}] \ (n=1,...,N)$ represent the bounding box for the $n$-th object. $x_{n}$ and $y_{n}$ are the coordinates of the center point in the box, and $w_{n}$ and $h_{n}$ are the width and height of the bounding box. We can use MLPs (multi-layer perceptrons) to predict object bounding boxes $\{\mathbf{b}_{1},...,\mathbf{b}_{N}\}$ from object embeddings $\{\mathbf{o}^{cr}_{1},...,\mathbf{o}^{cr}_{N}\}$. 
		
		\textbf{SGOR.} Inspired by prior fixed-set object detection works \cite{zhu2021deformable}, we build an SGOR mechanism. On the one hand, iterative offset regression generates more accurate bounding boxes than one-step prediction \cite{zhu2021deformable}. More importantly, we leverage SGOR to allow mutual enhancement between scene graph and object localization.
		
		Concretely, we first initialize $N$ object bounding boxes before the decoding in Sec.~\ref{Decoder}. Here, we use the same initialization as in \cite{zhu2021deformable}, which is based on deformable attention predictions. Then, in each block in our decoder, we leverage object boxes to enhance scene graphs by Eqn.~(\ref{eqn_sg}). After each block, new object embeddings $\{\mathbf{o}^{cr}_{l, 1},...,\mathbf{o}^{cr}_{l, N}\}$ are generated, where $l=1,..,L$ means the $l$-th decoder block. Our object embeddings include scene graph cues, and we leverage them to refine bounding boxes as follows:
		\begin{equation} 
			\mathbf{\Delta b}_{l,n} = MLP([\mathbf{o}_{l,n}^{cr}, \mathbf{b}_{l,n}])
		\end{equation}
		where $\mathbf{b}_{l,n} \in \mathbb{R}^{4}$ is the bounding box of the $n$-th object in the $l$-th decoder block. We concatenate the embedding $\mathbf{o}_{l,n}^{cr}$ and the bounding box $\mathbf{b}_{l,n}$ of this object. After that, an MLP with two linear layers and Sigmoid activation functions is used to predict the box offset $\mathbf{\Delta b}_{l,n} \in \mathbb{R}^{4}$. The bounding box of this object is refined as
		\begin{equation} 
			\mathbf{b}_{l+1,n} = \mathbf{b}_{l,n} + \mathbf{\Delta b}_{l,n}
		\end{equation}
		where $\mathbf{b}_{l+1,n}$ is the refined box and can be used as the bounding box in the next decoder block. For each object, the final bounding box prediction is the refined box $\mathbf{b}_{L+1, n} = \mathbf{b}_{L,n} + \mathbf{\Delta b}_{L,n}$ after the last decoder block.
		
		\textbf{Object category prediction.} We predict object and relation categories based on the object and predicate embeddings $\{\mathbf{o}^{cr}_{1}, ..., \mathbf{o}^{cr}_{N}, \mathbf{p}^{cr}\}$ in the final decoder block. Let $\mathbf{O} \in \mathbb{R}^{N \times D}$ be a matirx composed by object embeddings, i.e., $\mathbf{O} = \{\mathbf{o}^{cr}_{1}, ..., \mathbf{o}^{cr}_{N}\}$. To predict object categories, we first use two-layer MLPs to refine object embeddings $\mathbf{O}$ into $\mathbf{\widetilde{O}} \in \mathbb{R}^{N \times D}$. Then, we generate a similarity matrix between object and category embeddings:
		\begin{equation} 
			\mathbf{S}^{o} = \mathbf{\widetilde{O}}(\mathbf{F}^{o})^{T}
		\end{equation}
		where $\mathbf{S}^{o} \in \mathbb{R}^{N \times (C+1)}$ is the object category matrix. In $\mathbf{S}^{o}$, each element $s^{o}_{n,c}$ means the similarity between the $n$-th object and $c$-th object category. For each object, we can find the category with the highest similarity as its classification result, and it is a false proposal if its category is `\emph{no object}'.
		
		\textbf{Relation category prediction and joint learning.} For relation prediction, we first leverage Eqn.~(\ref{eqn_sg}) to generate scene graph embeddings $\mathbf{g}^{p}_{i,j}$ from our final object embeddings $\{\mathbf{o}^{cr}_{1}, ..., \mathbf{o}^{cr}_{N}\}$, predicate embedding $\mathbf{p}^{cr}$ and bounding boxes $\{\mathbf{b}_{L+1, 1}, ..., \mathbf{b}_{L+1, N}\}$. These scene graph embeddings compose a scene graph matrix $\mathbf{G}^{p} \in \mathbb{R}^{N(N-1) \times (3D+8)}$. Two-layer MLPs are used to transform $\mathbf{G}^{p}$ into $\widetilde{\mathbf{G}^{p}} \in \mathbb{R}^{N(N-1) \times D}$. A relation similarity matrix $\mathbf{S}^{p} \in \mathbb{R}^{N(N-1) \times (M+1)}$ is calculated as
		\begin{equation} \label{eqn_matching}
			\mathbf{S}^{p} = \widetilde{\mathbf{G}^{p}}(\mathbf{F}^{p})^{T}.
		\end{equation}
		Similar to object category prediction, we can determine the relation between every object pair by finding out the most similar relation category. There is no relation between two objects, if the predicted relation category is `\emph{no relation}'. In this way, we can generate OV relation classification results. Moreover, since relation categories are predicted from object embeddings and bounding boxes, we can obtain better object embeddings and object localization results by joint learning with relation training data.
		
		\textbf{Cross-modal learning.} We propose cross-modal learning to further exploit scene graph cues to classify OV objects. As OV detection models formulate object classification as a VL matching problem, the key to accurate classification is to learn consistent object and category embeddings. Therefore, we leverage relation supervision to enhance the consistency between object and category embeddings. Specifically, during training, we replace object embeddings in the scene graph matrix $\mathbf{G}^{p}$ with ground truth object category embeddings. Let $\mathbf{G}^{t} \in \mathbb{R}^{N(N-1) \times (3D+8)}$ represent the replaced matrix. We then use the same two-layer MLPs as in relation category prediction to transform the matrix into $\widetilde{\mathbf{G}^{t}} \in \mathbb{R}^{N(N-1) \times D}$, and generate the relation similarity matrix $\mathbf{S}^{t} \in \mathbb{R}^{N(N-1) \times (M+1)}$ as
		\begin{equation} \label{eqn_matching}
			\mathbf{S}^{t} = \widetilde{\mathbf{G}^{t}}(\mathbf{F}^{p})^{T}.
		\end{equation}
		By learning $\mathbf{S}^{t}$ with relation supervisions, $\mathbf{S}^{t}$ and $\mathbf{S}^{p}$ will be closed, and thus our object and category embeddings will be more consistent. Note that we only use cross-modal learning during the training stage, because we do not have ground truth object categories during inference.
		
		\subsection{Training} \label{Training}
		Our training objective contains four parts as follows:
		\begin{equation}
			Loss  = \lambda_{1}L_{bb} + \lambda_{2}L_{ocls} + \lambda_{3}L_{pcls} + \lambda_{4}L_{cml},
		\end{equation}
		where $L_{bb}$, $L_{ocls}$, $L_{pcls}$ and $L_{cml}$ are loss functions for object bounding box prediction, object category prediction, relation category prediction as well as cross-modal learning, respectively. $\lambda_{1}, \lambda_{2}, \lambda_{3}$ and $\lambda_{4}$ are hyperparameters to weight different losses.
		
		Our bounding box loss $L_{bb}$ is \emph{smooth l1} loss. Since SGOR predicts offsets and refines bounding boxes in every decoder block, we calculate a \emph{smooth l1} loss in each block, and the final bounding box loss $L_{bb}$ is their sum. Bipartite Hungarian matching is used to align predicted boxes with ground truths.
		
		Similar to previous OV works \cite{li2022grounded}, our object category loss $L_{ocls}$ is the sum of \emph{binary cross-entropy} losses for every element $s^{o}_{n,c}$ in our object similarity matrix $\mathbf{S}^{o}$. In particular, for each element $s^{o}_{n,c}$, we calculate a \emph{binary cross-entropy} loss between it and the ground truth. The ground truth is 1 when the $n$-th object belongs to the $c$-th category; otherwise, it is 0. Similarly, relation category loss $L_{pcls}$ and cross-modal leanring loss $L_{cml}$ are also \emph{binary cross-entropy} losses for similarity matrices $\mathbf{S}^{p}$ and $\mathbf{S}^{t}$, respectively.
		
		We leverage referring grounding data for training. A referring grounding sample contains an image, a language
		and bounding box annotations for every noun. The relation between each object pair can be extracted by language parsing tools, as described in Sec.~\ref{Encoding}. We take them as relation classification ground truths.
		Fixed-set object detection data can also be used during training. We only use $L_{bb}$ and $L_{ocls}$ for these data, while fixing the predicate token and relation prediction parts.
		
		\section{Experiments}
		\subsection{Experiment Settings}
		Following prior works \cite{lin2022learning, li2022grounded}, we evaluate the OV ability by zero-shot experiments on COCO \cite{lin2014microsoft} and LVIS \cite{gupta2019lvis}, and leverage VL data during training to recognize OV objects.
		
		\textbf{VL data.} Any extra VL can be used in the OV scenario. Here, we employ referring grounding data like previous methods \cite{li2022grounded}. Flickr30K Entities \cite{plummer2015flickr30k} includes 31K images with referring expressions and annotations. Visual Genome \cite{krishna2016visual} labels 108K image for referring grounding. We use these 140K training data. Our goal is to verify the effectiveness of our network rather than train a large pre-training model. Thus, we do not use millions of training data. Also, Visual Genome \cite{krishna2016visual} provides scene graph annotations, but we do not use them.
		
		\textbf{COCO.} The COCO 2017 dataset \cite{lin2014microsoft} contains 120K training and 5K validation images. We use the generalized zero-shot setting \cite{bansal2018zero}, which splits COCO into 48 base classes for training and 17 novel classes for validation. We combine 120K COCO training images and 140K grounding data to train our model. There are overlapped images between COCO \cite{lin2014microsoft} and Visual Genome \cite{krishna2016visual}. We remove them and training samples that contain 17 novel classes. 
		
		\textbf{LVIS.} There are 100K training and 20K validation images on LVIS \cite{gupta2019lvis}. Categories in LVIS are divided into 405 frequent, 461 common and 337 rare classes. We combine 886 frequent and common classes for training, while using 337 rare categories for validation. 140K grounding and 100K LVIS training data are mixed during training, where rare classes are removed. Since LVIS \cite{gupta2019lvis} also requires mask predictions, we use the external fully convolutional head in \cite{zang2022open} to generate masks based on decoder embeddings. We also test this LVIS-trained model on COCO to verify the cross-dataset ability.
		
		\textbf{Metrics.} We adopt $AP50$ for COCO \cite{lin2014microsoft} zero-shot detection, and $mAP$ for LVIS \cite{gupta2019lvis} as well as the cross-dataset validation.
		
		\begin{table*}
			\centering
			\caption{Zero-shot results on COCO validation. Note that several methods \cite{feng2022promptdet, lin2022learning, rasheed2022bridging} know novel classes during training, which is not practical for the OV task. }
			\scalebox{0.8}{
				\begin{tabular}{l|c|c|c|ccc} 
					\toprule             
					Method & Backbone & VL Pre-training & Training & novel ($AP50$) & base ($AP50$) & all ($AP50$) \\
					\hline
					\midrule
					\emph{without novel classes:} & \ & \ & \ & \  & \ & \ \\
					OVR-CNN \cite{zareian2021open} & Res-50 & - & 240K (COCO Caption, COCO base) &22.8 & - & - \\
					ViLD \cite{gu2021open}         & Res-50 & 400M (CLIP) & 120K (COCO base) &27.6 & 59.5 & 51.3  \\
					XPM \cite{huynh2022open}       & Res-50 & - & 5.1M (caption, OI, COCO base) & 29.9 & 46.3 & 42.0  \\
					OV-DETR \cite{zang2022open}   & Res-50 & 400M (CLIP) & 120K (COCO base) & 29.4 & \textbf{61.0} & 52.7 \\
					RegionCLIP \cite{zhong2022regionclip} & Res-50 & 400M (CLIP) & 3M (CC, COCO base) & 31.4 & 57.1 & 50.4  \\
					F-VLM \cite{kuo2022f} & Res-50 & 400M (CLIP) & 120K (COCO base) & 28.0 & - & 39.6 \\
					GLIP \cite{li2022grounded} (retrain) & Res-50 & - & 260K (grounding, COCO base) & 30.7 & 54.9 & 48.6 \\
					SGDN (Ours) & Res-50 & - & 260K (grounding, COCO base) & \textbf{37.5} & \textbf{61.0} & \textbf{54.9} \\
					\hline
					\emph{with novel classes:} & \ & \ & \ & \  & \ & \ \\
					PromptDet \cite{feng2022promptdet} & Res-50 & 400M (CLIP) & 400M (LAION, COCO) & 26.6 & - & 50.6  \\
					VLDet \cite{lin2022learning} & Res-50 & 400M (CLIP) & 240K (COCO Caption, COCO) & 32.0 & 50.6 & 45.8 \\
					Rasheed \emph{et al.} \cite{rasheed2022bridging} & Res-50 & 400M (CLIP, MViT) & 240K (COCO Caption, COCO) & 36.6 & 54.0 & 49.4 \\
					\bottomrule	
			\end{tabular}}
			\label{tab_result_coco_zs}
		\end{table*}		
		
		\begin{table*}
			\centering
			\caption{Zero-shot results on LVIS validation. Unlike COCO, $mAP$ on LVIS are calculated from mask results. We use the external head in \cite{zang2022open} to generate masks. Several methods \cite{feng2022promptdet, lin2022learning, rasheed2022bridging} add novel-class information to training, which is not practical from the OV perspective.}
			\scalebox{0.8}{
				\begin{tabular}{l|c|c|c|cccc} 
					\toprule             
					Method & Backbone & VL Pre-training & Training & rare ($mAP$) & common ($mAP$) & frequent ($mAP$) & all ($mAP$) \\
					\hline
					\midrule
					\emph{without novel classes:} & \ & \ & \ & \  & \ & \ & \ \\
					ViLD \cite{gu2021open}         & Res-50 & 400M (CLIP) & 100K (LVIS base) & 16.6 & 24.6 & 30.3 & 25.5 \\
					DetPro \cite{du2022learning}  & Res-50 & 400M (CLIP) & 100K (LVIS base) & 19.8 & 25.6 & 28.9 & 25.9  \\
					OV-DETR \cite{zang2022open}   & Res-50 & 400M (CLIP) & 100K (LVIS base) & 17.4 & 25.0 & 32.5 & 26.6 \\
					RegionCLIP \cite{zhong2022regionclip} & Res-50 & 400M (CLIP) & 3M (CC,LVIS base) & 17.1 & 27.4 & 34.0 & 28.2 \\
					F-VLM \cite{kuo2022f} & Res-50 & 400M (CLIP) & 100K (LVIS base) & 18.6 & - & - & 24.2\\
					GLIP \cite{li2022grounded} (retrain) & Res-50 & - & 240K (grounding, LVIS base) & 19.7 & 26.1 & 32.0 & 28.3 \\
					SGDN (Ours) & Res-50 & - & 240K (grounding, LVIS base) & \textbf{23.6} & 29.0 & \textbf{34.3} & \textbf{31.1} \\
					\hline
					\emph{with novel classes:} & \ & \ & \ & \  & \ & \ & \ \\
					PromptDet \cite{feng2022promptdet} & Res-50 & 400M (CLIP) & 400M (LAION, LVIS) & 21.4 & 23.3 & 29.3 & 25.3\\
					VLDet \cite{lin2022learning} & Res-50 & 400M (CLIP) & 3M (CC, LVIS) & 21.7 & \textbf{29.8} & \textbf{34.3} & 30.1\\
					Rasheed \emph{et al.} \cite{rasheed2022bridging} & Res-50 & 400M (CLIP, MViT) & 1.5M (ImageNet21K, LVIS) & 21.1 & 25.0 & 29.1 & 25.9\\
					\bottomrule	
			\end{tabular}}
			\label{tab_result_lvis_zs}
		\end{table*}
		
		\subsection{Implementation Details}
		We choose RoBERTa \cite{liu2019roberta} as our text encoder and add a linear layer to transform the textual feature dimension to $D$. $D$ is set to 512 in our experiments. We fix RoBERTa during training and only update the parameters of the linear layer. We do not use any prompt during training, only the prompt `A photo of a [query]' is used during inference. Our image encoder is Deformable DETR \cite{zhu2021deformable} with the ResNet50 \cite{he2016deep} backbone pre-trained on ImageNet. Deformable attention is also used in our decoder. The numbers $N$ and $L$ of object tokens and decoder blocks are set to 100 and 6, respectively. We train our model on two stages. VL data are used during the first-stage training, while fixed-set detection data is used in the second stage. $\lambda_{1}$, $\lambda_{2}$, $\lambda_{3}$ and $\lambda_{4}$ are simply set to 1.5, 1.5, 1.0 and 1.0, respectively, and fixed for all datasets. 
		Other network and training settings are the same as Deformable DETR \cite{zhu2021deformable}. All experiments are conducted on the Pytorch platform \cite{paszke2019pytorch} with 8 V100 GPUs.				
		
		\subsection{Main Results}
		We report the zero-shot results of our model and other state-of-the-art methods on COCO \cite{lin2014microsoft} in Table \ref{tab_result_coco_zs}. Our SGDN outperforms Rasheed \emph{et al.} \cite{rasheed2022bridging}, which achieve the best performance for novel classes in previous methods. Rasheed \emph{et al.} \cite{rasheed2022bridging} use the large-scale VL pretraining CLIP \cite{radford2021learning} and add COCO Caption data for training. They also leverage novel-class information during training, which is not practical for OV detection. In prior works without novel-class information, RegionCLIP \cite{zhong2022regionclip} shows the highest accuracy, which also uses CLIP \cite{radford2021learning} and three million extra data. Compared with it, our SGDN yields improvements of 6.1\% for novel classes. GLIP \cite{li2022grounded} also uses referring grounding training data, but it is trained with 27 million data as a VL pre-training. Rather than designing a pre-training, our work aims to provide an architecture for OV detection. Therefore, we reproduce GLIP \cite{li2022grounded} with our 260K training data for a fair comparison. Our SGDN outperforms it by 6.8\% for novel classes. We also achieve the best accuracy for all classes.
		
		Table \ref{tab_result_lvis_zs} shows the zero-shot results on LVIS \cite{gupta2019lvis}. We outperform the previous state-of-the-art method VLDet \cite{lin2022learning} by 1.9\% for rare classes. Note that, VLDet \cite{lin2022learning} also uses novel-class information during training. Compared with GLIP \cite{li2022grounded}, which uses the same VL training data, we achieve improvements of 3.9\% for rare classes and 2.8\% for all classes. 
		
		As several works \cite{gu2021open, du2022learning, zang2022open, cai2022x, li2022grounded} show cross-dataset results to further verify the OV ability, we also report these results in Table \ref{tab_result_cd}. Our method outperforms all these methods except the original GLIP \cite{li2022grounded}, which uses 27 million training data and a large Swin-L \cite{liu2021swin}. Compared with GLIP \cite{li2022grounded} with the same backbone and training data, our SGDN obtains gains of 3.6\%. Meanwhile, X-DETR \cite{cai2022x} also employs grounding data for training, and we outperform it by 14\%. All these superior results demonstrate the effectiveness of our scene-graph-based framework, as well as our proposed SGDecoder and SGPred modules. 

		\begin{figure*}
	\centerline{\includegraphics[scale=0.35]{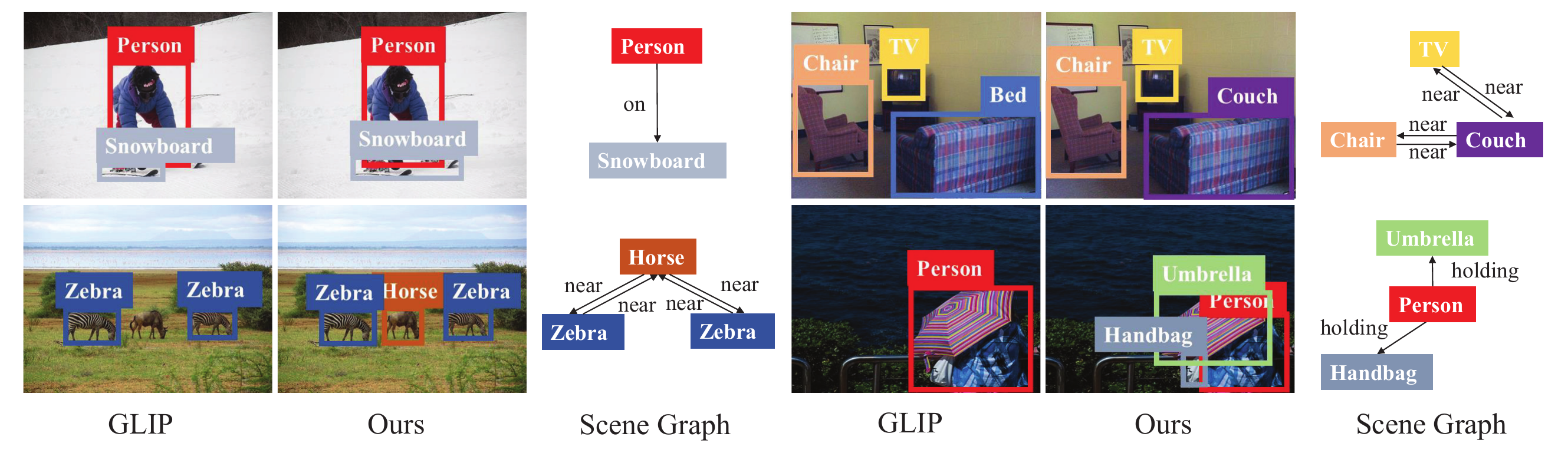}}
	\caption{Visualized zero-shot testing results on COCO validation. GLIP [18] is retrained on our data. We also show our scene graph results. Here, we use the 50 candidate relation categories defined in Visual Genome. GLIP [18] mislocalizes, misclassifies and misses some unseen objects, such as the `\emph{snowboard}', `\emph{couch}', `\emph{umbrella}' as well as `\emph{handbag}' objects, while our SGDN avoids these mistakes by exploiting scene graph cues.}
	\label{fig_coco}
\end{figure*}		
		
		\subsection{Ablation Study} 
		To further verify the effectiveness of our SGDN, we conduct ablation studies on zero-shot COCO. All models are trained with 260K VL and COCO base data, and use the ResNet-50 backbone.
		
		\textbf{Scene graph for OV detection.}  We report the effects of our main components in Table \ref{tab_main}. Since we use deformable attention \cite{zhu2021deformable}, we first build a Deformable-DETR-based OV detection model, `Model A', as our baseline. In `Model A', we add a text encoder to Deformable DETR \cite{zhu2021deformable} and replace its classification head with our object category prediction. 
		Then, we design a scene-graph-based `Model B', where we incorporate the predicate token $\mathbf{p}$ as well as the relation category prediction module to `Model A', and extract scene graphs for training. For novel classes, `Model B' outperforms `Model A' and GLIP \cite{li2022grounded} by 2.4\% and 1.6\%, respectively. These results show the effectiveness of scene graphs for OV object detection. 
		
		\textbf{Main component.} Our SGDecoder (`Model C') outperforms `Model B' by 2.3\% for novel classes, because our SGDecoder with SSGA leverages scene graphs to better embed objects. Our SGPred (`Model D') achieves gains of 3.5\% and 3.2\% for novel and all classes, which demonstrates the effectiveness of our SGOR and cross-modal learning.
		Compared with `Model B', our final SGDN yields improvements of 5.2\% and 4.6\% for novel and all classes, respectively.
		
		\begin{table*}
			\centering
			\caption{Cross-dataset results on COCO validation. We use the model trained on LVIS. }
			\scalebox{0.8}{
				\begin{tabular}{l|c|c|c|c}
					\toprule
					Method & Backbone & VL Pre-training & Training & $mAP$  \\
					\hline
					\midrule
					ViLD \cite{gu2021open}     & Res-50  & 400M (CLIP) & 100K (LVIS base) & 36.6 \\
					DetPro \cite{du2022learning} & Res-50 & 400M (CLIP) & 100K (LVIS base) & 34.9 \\
					OV-DETR \cite{zang2022open} & Res-50 & 400M (CLIP) & 100K (LVIS base) & 38.1  \\
					X-DETR \cite{cai2022x}      & Res-101 & -  & 14M (grounding, detection, caption) & 26.5 \\
					GLIP \cite{li2022grounded}    & Swin-L & - & 27M (grounding, detection, caption) & \textbf{49.8} \\
					GLIP \cite{li2022grounded} (retrain) & Res-50 & - & 240K (grounding, LVIS base) & 36.9 \\
					SGDN (Ours) & Res-50 & - & 240K (grounding, LVIS base)     & 40.5  \\				
					\bottomrule	
			\end{tabular}}
			\label{tab_result_cd}
		\end{table*}

		\begin{table}
			\centering
			\caption{The effects of main parts in our method on COCO zero-shot validation. `SG' means scene graph information. }
			\scalebox{0.8}{
				\begin{tabular}{l|ccc|cc} 
					\toprule 
					&    &           &        & \multicolumn{2}{c}{$AP50$} \\
					Model & SG & SGDecoder & SGPred & novel & all \\
					\midrule
					GLIP \cite{li2022grounded} (retrain) & & & & 30.7 & 48.6 \\
					\hline
					A: OV Deformable DETR & & & & 29.9 & 48.9 \\
					B: OV scene graph model & \checkmark & & & 32.3 & 50.3 \\
					C: SGDecoder-based model & \checkmark & \checkmark & & 34.6 & 51.4  \\
					D: SGPred-based model & \checkmark &  & \checkmark & 35.8 & 53.5 \\
					SGDN & \checkmark & \checkmark & \checkmark & 37.5 & 54.9 \\
					\bottomrule	
			\end{tabular}}
			\label{tab_main}
		\end{table}

		\textbf{Dissecting SGDecoder.} We then dissect our SGDecoder in Table~\ref{tab_SGDecoder}. If we remove SSGA from `Model C', the model is equal to `Model B' and the performance significantly decreases. In `Model C w/o box', we use SSGA but remove bounding boxes from scene graph embeddings. Compare to `Model B', this model achieves improvements of 1.5\% for novel classes. The reason is that SSGA better exploits scene graph information for OV detection. Bounding boxes generate gains of 0.8\% for novel classes. In `Model C w/o sparse', we employ vanilla deformable attention instead of the sparse one. The performance only slightly increases. However, vanilla attention requires much more computational costs than sparse attention.

		\begin{table}
			\centering
			\caption{The effects of different settings in our SGPred on COCO zero-shot validation. `W/o' is without. We do not use SGDecoder in this experiment.}
			\scalebox{0.8}{
				\begin{tabular}{l|cc} 
					\toprule 
					Model & novel ($AP50$) & all ($AP50$) \\
					\midrule
					D: SGPred-based model & 35.8 & 53.5 \\
					w/o SGOR & 34.3 & 51.2  \\
					w/o cross-modal learning & 34.7 & 52.6  \\
					\bottomrule	
			\end{tabular}}
			\label{tab_SGPred}
		\end{table}

		\textbf{Dissecting SGPred.} In Table~\ref{tab_SGPred}, we show the effects of main parts in our SGPred. Our SGOR mutually improves object localization and scene graph embedding, and thus yields gains of 1.5\% and 2.3\% for novel and all classes. Cross-modal learning increases the performance for novel classes by 1.1\%, benefiting from the scene-graph-based cross-modal consistency enhancement.
		
		\begin{table}
			\centering
			\caption{The effects of different settings in SGDecoder on COCO zero-shot validation. `W/o' means without. SGPred is not used in this experiment.}
			\scalebox{0.8}{
				\begin{tabular}{l|cc} 
					\toprule 
					Model & novel ($AP50$) & all ($AP50$) \\
					\midrule
					C: SGDecoder-based model & 34.6 & 51.4 \\
					w/o SSGA (Model B) & 32.3 & 50.3 \\
					w/o box & 33.8 & 51.1  \\
					w/o sparse & 34.9 & 51.2  \\
					\bottomrule	
			\end{tabular}}
			\label{tab_SGDecoder}
		\end{table}
		
		We show our OV SGDet ability in Table \ref{tab_OVSG}. We conduct this experiment on Visual Genome \cite{krishna2016visual}, and use the dataset split provided by \cite{he2022towards}. The training set includes 70\% seen object and relation classes in Visual Genome grounding and scene graph data, while the 30\% unseen object and relation classes are used for validation. Metrics are $Recall@50$ and $Recall@100$. It can be seen that our SGDN significantly outperforms previous OV scene graph methods on the OV SGCls task and can also predict OV SGDet results.

	\begin{figure*}
	\centerline{\includegraphics[scale=0.35]{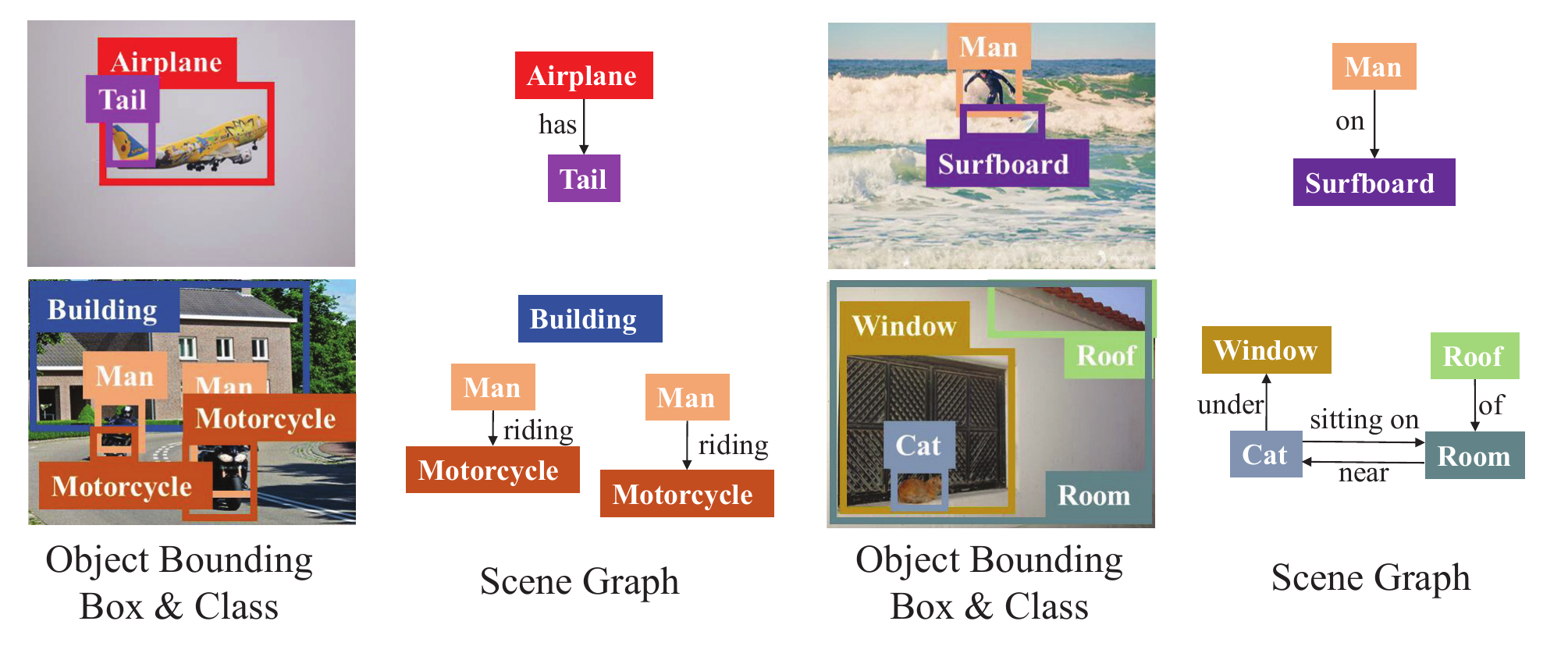}}
	\caption{Visualized OV scene graph detection results of our SGDN on Visual Genome validation. Our method can tackle this task, and predicts bounding boxes, object and relation classes.}
	\label{fig_vg}
\end{figure*}
		
		\begin{table}
			\centering
			\caption{OV scene graph generation results on Visual Genome.}
			\scalebox{0.8}{
				\begin{tabular}{l|cc|cc} 
					\toprule 
					&  \multicolumn{2}{c|}{OV SGCls} & \multicolumn{2}{c}{OV SGDet} \\
					Model & $R@50$ & $R@100$   & $R@50$ & $R@100$ \\
					\midrule
					SVRP \cite{he2022towards} & 19.1 & 21.5 & - & - \\
					SGDN (Ours) & 24.8 & 30.2 & 9.8 &  14.5 \\
					\bottomrule	
			\end{tabular}}
			\label{tab_OVSG}
		\end{table}	

		\textbf{Qualitative results.} Fig. \ref{fig_coco} shows qualitative results on COCO. It can be observed that GLIP \cite{li2022grounded} misclassifies some unseen objects, such as the `\emph{couch}' in the upper right image in Fig. \ref{fig_coco}. We exploit scene graph information to better recognize OV objects. Normally, the object `\emph{near}' a `\emph{TV}' and a `\emph{chair}' is more like a `\emph{couch}' than a `\emph{bed}'. Therefore, our approach reduces this classification error. GLIP [18] also misses a number of unseen objects. For example, in the bottom left and right images in Fig. \ref{fig_coco}, the `\emph{horse}', `\emph{umbrella}' and `\emph{handbag}' are missed by GLIP \cite{li2022grounded}. Our SGDN successfully discovers them by exploiting scene graph cues `\emph{zebra near}' and `\emph{person holding}'. Moreover, SGDN can better localize unseen objects (e.g., the `\emph{snowboard}' in the upper left image in Fig. \ref{fig_coco})  based on scene graphs. These results demonstrate the effectiveness of our SGDN for OV object discovery, classification and localization.

        \textbf{OV scene graph detection.} Scene graph generation contains three tasks. The simplest is predicate classification (PredCls), where object bounding boxes and classes are provided, and only object relations (predicates) need to be classified. The second is scene graph classification (SGCls). By given object bounding boxes, SGCls expects to classify objects and relations. The hardest is scene graph detection (SGDet), which requires to predict all bounding boxes, object and relation classes. Previous OV scene graph generation methods cannot deal with the OV SGDet task, because they are not able to detect OV objects \cite{he2022towards}. Different from them, our SGDN can simultaneously detect OV objects and relations, and thus generates OV SGDet predictions.

     	We visualize OV scene graph detection results in Fig. \ref{fig_vg}. Since SVPR \cite{he2022towards} does not release its source code, we only show the results from our SGDN in Fig. \ref{fig_vg}. We successfully localize and classify unseen objects, e.g., `\emph{man}', `\emph{cat}' and `\emph{window}'. Meanwhile, unseen relations such as `\emph{sitting on}' are also predicted by our SGDN.
		
		\section{Conclusion}
		In this paper, we have presented SGDN, a scene-graph-based network for OV object detection. We first introduce an SGDecoder to generate object and relation embeddings, where an SSGA module is presented to leverage scene-graph cues for OV object discovery, classification and localization. Secondly, an SGPred method is designed to predict OV object detection and scene graph results, including SGOR and cross-modal learning. In SGOR, scene graphs and object localization are iteratively improved by each other. Cross-modal learning takes scene graphs as bridges to enhance the consistency between cross-modal embeddings for OV object classification. Extensive experiments on two OV detection datasets demonstrate the effectiveness of our SGDN. We also show the OV scene graph detection ability of SGDN, which cannot be solved by previous OV scene graph generation approaches.

		\bibliographystyle{ACM-Reference-Format}
		\bibliography{my_reference}


\begin{thebibliography}{54}


\ifx \showCODEN    \undefined \def \showCODEN     #1{\unskip}     \fi
\ifx \showDOI      \undefined \def \showDOI       #1{#1}\fi
\ifx \showISBNx    \undefined \def \showISBNx     #1{\unskip}     \fi
\ifx \showISBNxiii \undefined \def \showISBNxiii  #1{\unskip}     \fi
\ifx \showISSN     \undefined \def \showISSN      #1{\unskip}     \fi
\ifx \showLCCN     \undefined \def \showLCCN      #1{\unskip}     \fi
\ifx \shownote     \undefined \def \shownote      #1{#1}          \fi
\ifx \showarticletitle \undefined \def \showarticletitle #1{#1}   \fi
\ifx \showURL      \undefined \def \showURL       {\relax}        \fi
\providecommand\bibfield[2]{#2}
\providecommand\bibinfo[2]{#2}
\providecommand\natexlab[1]{#1}
\providecommand\showeprint[2][]{arXiv:#2}

\bibitem[Bansal et~al\mbox{.}(2018)]%
        {bansal2018zero}
\bibfield{author}{\bibinfo{person}{Ankan Bansal}, \bibinfo{person}{Karan
  Sikka}, \bibinfo{person}{Gaurav Sharma}, \bibinfo{person}{Rama Chellappa},
  {and} \bibinfo{person}{Ajay Divakaran}.} \bibinfo{year}{2018}\natexlab{}.
\newblock \showarticletitle{Zero-shot object detection}. In
  \bibinfo{booktitle}{\emph{Proceedings of the European Conference on Computer
  Vision (ECCV)}}. \bibinfo{pages}{384--400}.
\newblock


\bibitem[Cai et~al\mbox{.}(2022)]%
        {cai2022x}
\bibfield{author}{\bibinfo{person}{Zhaowei Cai}, \bibinfo{person}{Gukyeong
  Kwon}, \bibinfo{person}{Avinash Ravichandran}, \bibinfo{person}{Erhan Bas},
  \bibinfo{person}{Zhuowen Tu}, \bibinfo{person}{Rahul Bhotika}, {and}
  \bibinfo{person}{Stefano Soatto}.} \bibinfo{year}{2022}\natexlab{}.
\newblock \showarticletitle{X-DETR: A Versatile Architecture for Instance-wise
  Vision-Language Tasks}.
\newblock \bibinfo{journal}{\emph{European Conference on Computer Vision}}
  (\bibinfo{year}{2022}).
\newblock


\bibitem[Cao et~al\mbox{.}(2022)]%
        {cao2022correspondence}
\bibfield{author}{\bibinfo{person}{Meng Cao}, \bibinfo{person}{Ji Jiang},
  \bibinfo{person}{Long Chen}, {and} \bibinfo{person}{Yuexian Zou}.}
  \bibinfo{year}{2022}\natexlab{}.
\newblock \showarticletitle{Correspondence matters for video referring
  expression comprehension}. In \bibinfo{booktitle}{\emph{Proceedings of the
  30th ACM International Conference on Multimedia}}.
  \bibinfo{pages}{4967--4976}.
\newblock


\bibitem[Chen(2021)]%
        {chen2021towards}
\bibfield{author}{\bibinfo{person}{Shaoxiang Chen}.}
  \bibinfo{year}{2021}\natexlab{}.
\newblock \showarticletitle{Towards bridging video and language by caption
  generation and sentence localization}. In
  \bibinfo{booktitle}{\emph{Proceedings of the 29th ACM International
  Conference on Multimedia}}. \bibinfo{pages}{2964--2968}.
\newblock


\bibitem[Du et~al\mbox{.}(2022)]%
        {du2022learning}
\bibfield{author}{\bibinfo{person}{Yu Du}, \bibinfo{person}{Fangyun Wei},
  \bibinfo{person}{Zihe Zhang}, \bibinfo{person}{Miaojing Shi},
  \bibinfo{person}{Yue Gao}, {and} \bibinfo{person}{Guoqi Li}.}
  \bibinfo{year}{2022}\natexlab{}.
\newblock \showarticletitle{Learning to Prompt for Open-Vocabulary Object
  Detection with Vision-Language Model}. In
  \bibinfo{booktitle}{\emph{Proceedings of the IEEE/CVF Conference on Computer
  Vision and Pattern Recognition}}. \bibinfo{pages}{14084--14093}.
\newblock


\bibitem[Feng et~al\mbox{.}(2022)]%
        {feng2022promptdet}
\bibfield{author}{\bibinfo{person}{Chengjian Feng}, \bibinfo{person}{Yujie
  Zhong}, \bibinfo{person}{Zequn Jie}, \bibinfo{person}{Xiangxiang Chu},
  \bibinfo{person}{Haibing Ren}, \bibinfo{person}{Xiaolin Wei},
  \bibinfo{person}{Weidi Xie}, {and} \bibinfo{person}{Lin Ma}.}
  \bibinfo{year}{2022}\natexlab{}.
\newblock \showarticletitle{Promptdet: Towards open-vocabulary detection using
  uncurated images}. In \bibinfo{booktitle}{\emph{European Conference on
  Computer Vision}}. \bibinfo{pages}{701--717}.
\newblock


\bibitem[Gao et~al\mbox{.}(2022)]%
        {gao2021towards}
\bibfield{author}{\bibinfo{person}{Mingfei Gao}, \bibinfo{person}{Chen Xing},
  \bibinfo{person}{Juan~Carlos Niebles}, \bibinfo{person}{Junnan Li},
  \bibinfo{person}{Ran Xu}, \bibinfo{person}{Wenhao Liu}, {and}
  \bibinfo{person}{Caiming Xiong}.} \bibinfo{year}{2022}\natexlab{}.
\newblock \showarticletitle{Open Vocabulary Object Detection with Pseudo
  Bounding-Box Labels}.
\newblock \bibinfo{journal}{\emph{European Conference on Computer Vision}}
  (\bibinfo{year}{2022}).
\newblock


\bibitem[Gu et~al\mbox{.}(2022)]%
        {gu2021open}
\bibfield{author}{\bibinfo{person}{Xiuye Gu}, \bibinfo{person}{Tsung-Yi Lin},
  \bibinfo{person}{Weicheng Kuo}, {and} \bibinfo{person}{Yin Cui}.}
  \bibinfo{year}{2022}\natexlab{}.
\newblock \showarticletitle{Open-vocabulary Object Detection via Vision and
  Language Knowledge Distillation}. In \bibinfo{booktitle}{\emph{International
  Conference on Learning Representations}}.
\newblock


\bibitem[Gupta et~al\mbox{.}(2019)]%
        {gupta2019lvis}
\bibfield{author}{\bibinfo{person}{Agrim Gupta}, \bibinfo{person}{Piotr
  Dollar}, {and} \bibinfo{person}{Ross Girshick}.}
  \bibinfo{year}{2019}\natexlab{}.
\newblock \showarticletitle{Lvis: A dataset for large vocabulary instance
  segmentation}. In \bibinfo{booktitle}{\emph{Proceedings of the IEEE/CVF
  conference on computer vision and pattern recognition}}.
  \bibinfo{pages}{5356--5364}.
\newblock


\bibitem[He et~al\mbox{.}(2016)]%
        {he2016deep}
\bibfield{author}{\bibinfo{person}{Kaiming He}, \bibinfo{person}{Xiangyu
  Zhang}, \bibinfo{person}{Shaoqing Ren}, {and} \bibinfo{person}{Jian Sun}.}
  \bibinfo{year}{2016}\natexlab{}.
\newblock \showarticletitle{Deep residual learning for image recognition}. In
  \bibinfo{booktitle}{\emph{Proceedings of the IEEE Conference on Computer
  Vision and Pattern Recognition}}. \bibinfo{pages}{770--778}.
\newblock


\bibitem[He et~al\mbox{.}(2022)]%
        {he2022towards}
\bibfield{author}{\bibinfo{person}{Tao He}, \bibinfo{person}{Lianli Gao},
  \bibinfo{person}{Jingkuan Song}, {and} \bibinfo{person}{Yuan-Fang Li}.}
  \bibinfo{year}{2022}\natexlab{}.
\newblock \showarticletitle{Towards open-vocabulary scene graph generation with
  prompt-based finetuning}. In \bibinfo{booktitle}{\emph{European Conference on
  Computer Vision}}. \bibinfo{pages}{56--73}.
\newblock


\bibitem[Huynh et~al\mbox{.}(2022)]%
        {huynh2022open}
\bibfield{author}{\bibinfo{person}{Dat Huynh}, \bibinfo{person}{Jason Kuen},
  \bibinfo{person}{Zhe Lin}, \bibinfo{person}{Jiuxiang Gu}, {and}
  \bibinfo{person}{Ehsan Elhamifar}.} \bibinfo{year}{2022}\natexlab{}.
\newblock \showarticletitle{Open-vocabulary instance segmentation via robust
  cross-modal pseudo-labeling}. In \bibinfo{booktitle}{\emph{Proceedings of the
  IEEE/CVF Conference on Computer Vision and Pattern Recognition}}.
  \bibinfo{pages}{7020--7031}.
\newblock


\bibitem[Kenton and Toutanova(2019)]%
        {kenton2019bert}
\bibfield{author}{\bibinfo{person}{Jacob Devlin Ming-Wei~Chang Kenton} {and}
  \bibinfo{person}{Lee~Kristina Toutanova}.} \bibinfo{year}{2019}\natexlab{}.
\newblock \showarticletitle{BERT: Pre-training of Deep Bidirectional
  Transformers for Language Understanding}. In
  \bibinfo{booktitle}{\emph{Proceedings of NAACL-HLT}}.
  \bibinfo{pages}{4171--4186}.
\newblock


\bibitem[Krishna et~al\mbox{.}(2016)]%
        {krishna2016visual}
\bibfield{author}{\bibinfo{person}{Ranjay Krishna}, \bibinfo{person}{Yuke Zhu},
  \bibinfo{person}{Oliver Groth}, \bibinfo{person}{Justin Johnson},
  \bibinfo{person}{Kenji Hata}, \bibinfo{person}{Joshua Kravitz},
  \bibinfo{person}{Stephanie Chen}, \bibinfo{person}{Yannis Kalantidis},
  \bibinfo{person}{Li-Jia Li}, \bibinfo{person}{David~A Shamma},
  {et~al\mbox{.}}} \bibinfo{year}{2016}\natexlab{}.
\newblock \showarticletitle{Visual genome: Connecting language and vision using
  crowdsourced dense image annotations}.
\newblock \bibinfo{journal}{\emph{IJCV}} (\bibinfo{year}{2016}).
\newblock


\bibitem[Kuo et~al\mbox{.}(2022)]%
        {kuo2022findit}
\bibfield{author}{\bibinfo{person}{Weicheng Kuo}, \bibinfo{person}{Fred
  Bertsch}, \bibinfo{person}{Wei Li}, \bibinfo{person}{AJ Piergiovanni},
  \bibinfo{person}{Mohammad Saffar}, {and} \bibinfo{person}{Anelia Angelova}.}
  \bibinfo{year}{2022}\natexlab{}.
\newblock \showarticletitle{FindIt: Generalized Localization with Natural
  Language Queries}.
\newblock \bibinfo{journal}{\emph{European Conference on Computer Vision}}
  (\bibinfo{year}{2022}).
\newblock


\bibitem[Kuo et~al\mbox{.}(2023)]%
        {kuo2022f}
\bibfield{author}{\bibinfo{person}{Weicheng Kuo}, \bibinfo{person}{Yin Cui},
  \bibinfo{person}{Xiuye Gu}, \bibinfo{person}{AJ Piergiovanni}, {and}
  \bibinfo{person}{Anelia Angelova}.} \bibinfo{year}{2023}\natexlab{}.
\newblock \showarticletitle{F-VLM: Open-Vocabulary Object Detection upon Frozen
  Vision and Language Models}.
\newblock \bibinfo{journal}{\emph{International Conference on Learning
  Representations}} (\bibinfo{year}{2023}).
\newblock


\bibitem[Li et~al\mbox{.}(2021a)]%
        {li2021bottom}
\bibfield{author}{\bibinfo{person}{Liuwu Li}, \bibinfo{person}{Yuqi Bu}, {and}
  \bibinfo{person}{Yi Cai}.} \bibinfo{year}{2021}\natexlab{a}.
\newblock \showarticletitle{Bottom-Up and Bidirectional Alignment for Referring
  Expression Comprehension}. In \bibinfo{booktitle}{\emph{Proceedings of the
  29th ACM International Conference on Multimedia}}.
  \bibinfo{pages}{5167--5175}.
\newblock


\bibitem[Li et~al\mbox{.}(2022c)]%
        {li2022grounded}
\bibfield{author}{\bibinfo{person}{Liunian~Harold Li},
  \bibinfo{person}{Pengchuan Zhang}, \bibinfo{person}{Haotian Zhang},
  \bibinfo{person}{Jianwei Yang}, \bibinfo{person}{Chunyuan Li},
  \bibinfo{person}{Yiwu Zhong}, \bibinfo{person}{Lijuan Wang},
  \bibinfo{person}{Lu Yuan}, \bibinfo{person}{Lei Zhang},
  \bibinfo{person}{Jenq-Neng Hwang}, {et~al\mbox{.}}}
  \bibinfo{year}{2022}\natexlab{c}.
\newblock \showarticletitle{Grounded language-image pre-training}. In
  \bibinfo{booktitle}{\emph{Proceedings of the IEEE/CVF Conference on Computer
  Vision and Pattern Recognition}}. \bibinfo{pages}{10965--10975}.
\newblock


\bibitem[Li et~al\mbox{.}(2022b)]%
        {li2022sgtr}
\bibfield{author}{\bibinfo{person}{Rongjie Li}, \bibinfo{person}{Songyang
  Zhang}, {and} \bibinfo{person}{Xuming He}.} \bibinfo{year}{2022}\natexlab{b}.
\newblock \showarticletitle{Sgtr: End-to-end scene graph generation with
  transformer}. In \bibinfo{booktitle}{\emph{Proceedings of the IEEE/CVF
  Conference on Computer Vision and Pattern Recognition}}.
  \bibinfo{pages}{19486--19496}.
\newblock


\bibitem[Li et~al\mbox{.}(2021b)]%
        {li2021bipartite}
\bibfield{author}{\bibinfo{person}{Rongjie Li}, \bibinfo{person}{Songyang
  Zhang}, \bibinfo{person}{Bo Wan}, {and} \bibinfo{person}{Xuming He}.}
  \bibinfo{year}{2021}\natexlab{b}.
\newblock \showarticletitle{Bipartite graph network with adaptive message
  passing for unbiased scene graph generation}. In
  \bibinfo{booktitle}{\emph{Proceedings of the IEEE/CVF Conference on Computer
  Vision and Pattern Recognition}}. \bibinfo{pages}{11109--11119}.
\newblock


\bibitem[Li et~al\mbox{.}(2022a)]%
        {li2022integrating}
\bibfield{author}{\bibinfo{person}{Xingchen Li}, \bibinfo{person}{Long Chen},
  \bibinfo{person}{Wenbo Ma}, \bibinfo{person}{Yi Yang}, {and}
  \bibinfo{person}{Jun Xiao}.} \bibinfo{year}{2022}\natexlab{a}.
\newblock \showarticletitle{Integrating object-aware and interaction-aware
  knowledge for weakly supervised scene graph generation}. In
  \bibinfo{booktitle}{\emph{Proceedings of the 30th ACM International
  Conference on Multimedia}}. \bibinfo{pages}{4204--4213}.
\newblock


\bibitem[Li et~al\mbox{.}(2017)]%
        {li2017scene}
\bibfield{author}{\bibinfo{person}{Yikang Li}, \bibinfo{person}{Wanli Ouyang},
  \bibinfo{person}{Bolei Zhou}, \bibinfo{person}{Kun Wang}, {and}
  \bibinfo{person}{Xiaogang Wang}.} \bibinfo{year}{2017}\natexlab{}.
\newblock \showarticletitle{Scene graph generation from objects, phrases and
  region captions}. In \bibinfo{booktitle}{\emph{Proceedings of the IEEE
  international conference on computer vision}}. \bibinfo{pages}{1261--1270}.
\newblock


\bibitem[Lin et~al\mbox{.}(2023)]%
        {lin2022learning}
\bibfield{author}{\bibinfo{person}{Chuang Lin}, \bibinfo{person}{Peize Sun},
  \bibinfo{person}{Yi Jiang}, \bibinfo{person}{Ping Luo},
  \bibinfo{person}{Lizhen Qu}, \bibinfo{person}{Gholamreza Haffari},
  \bibinfo{person}{Zehuan Yuan}, {and} \bibinfo{person}{Jianfei Cai}.}
  \bibinfo{year}{2023}\natexlab{}.
\newblock \showarticletitle{Learning Object-Language Alignments for
  Open-Vocabulary Object Detection}.
\newblock \bibinfo{journal}{\emph{International Conference on Learning
  Representations}} (\bibinfo{year}{2023}).
\newblock


\bibitem[Lin et~al\mbox{.}(2014)]%
        {lin2014microsoft}
\bibfield{author}{\bibinfo{person}{Tsung-Yi Lin}, \bibinfo{person}{Michael
  Maire}, \bibinfo{person}{Serge Belongie}, \bibinfo{person}{James Hays},
  \bibinfo{person}{Pietro Perona}, \bibinfo{person}{Deva Ramanan},
  \bibinfo{person}{Piotr Doll{\'a}r}, {and} \bibinfo{person}{C~Lawrence
  Zitnick}.} \bibinfo{year}{2014}\natexlab{}.
\newblock \showarticletitle{Microsoft coco: Common objects in context}. In
  \bibinfo{booktitle}{\emph{Proceedings of the European Conference on Computer
  Vision}}. Springer, \bibinfo{pages}{740--755}.
\newblock


\bibitem[Liu et~al\mbox{.}(2019)]%
        {liu2019roberta}
\bibfield{author}{\bibinfo{person}{Yinhan Liu}, \bibinfo{person}{Myle Ott},
  \bibinfo{person}{Naman Goyal}, \bibinfo{person}{Jingfei Du},
  \bibinfo{person}{Mandar Joshi}, \bibinfo{person}{Danqi Chen},
  \bibinfo{person}{Omer Levy}, \bibinfo{person}{Mike Lewis},
  \bibinfo{person}{Luke Zettlemoyer}, {and} \bibinfo{person}{Veselin
  Stoyanov}.} \bibinfo{year}{2019}\natexlab{}.
\newblock \showarticletitle{Roberta: A robustly optimized bert pretraining
  approach}.
\newblock \bibinfo{journal}{\emph{arXiv preprint arXiv:1907.11692}}
  (\bibinfo{year}{2019}).
\newblock


\bibitem[Liu et~al\mbox{.}(2018)]%
        {liu2018structure}
\bibfield{author}{\bibinfo{person}{Yong Liu}, \bibinfo{person}{Ruiping Wang},
  \bibinfo{person}{Shiguang Shan}, {and} \bibinfo{person}{Xilin Chen}.}
  \bibinfo{year}{2018}\natexlab{}.
\newblock \showarticletitle{Structure inference net: Object detection using
  scene-level context and instance-level relationships}. In
  \bibinfo{booktitle}{\emph{Proceedings of the IEEE conference on computer
  vision and pattern recognition}}. \bibinfo{pages}{6985--6994}.
\newblock


\bibitem[Liu et~al\mbox{.}(2021)]%
        {liu2021swin}
\bibfield{author}{\bibinfo{person}{Ze Liu}, \bibinfo{person}{Yutong Lin},
  \bibinfo{person}{Yue Cao}, \bibinfo{person}{Han Hu}, \bibinfo{person}{Yixuan
  Wei}, \bibinfo{person}{Zheng Zhang}, \bibinfo{person}{Stephen Lin}, {and}
  \bibinfo{person}{Baining Guo}.} \bibinfo{year}{2021}\natexlab{}.
\newblock \showarticletitle{Swin transformer: Hierarchical vision transformer
  using shifted windows}.
\newblock \bibinfo{journal}{\emph{ICCV}} (\bibinfo{year}{2021}).
\newblock


\bibitem[Lu et~al\mbox{.}(2016)]%
        {lu2016visual}
\bibfield{author}{\bibinfo{person}{Cewu Lu}, \bibinfo{person}{Ranjay Krishna},
  \bibinfo{person}{Michael Bernstein}, {and} \bibinfo{person}{Li Fei-Fei}.}
  \bibinfo{year}{2016}\natexlab{}.
\newblock \showarticletitle{Visual relationship detection with language
  priors}. In \bibinfo{booktitle}{\emph{Computer Vision--ECCV 2016: 14th
  European Conference, Amsterdam, The Netherlands, October 11--14, 2016,
  Proceedings, Part I 14}}. \bibinfo{pages}{852--869}.
\newblock


\bibitem[Lyu et~al\mbox{.}(2020)]%
        {lyu2020vtgraphnet}
\bibfield{author}{\bibinfo{person}{Fan Lyu}, \bibinfo{person}{Wei Feng}, {and}
  \bibinfo{person}{Song Wang}.} \bibinfo{year}{2020}\natexlab{}.
\newblock \showarticletitle{vtGraphNet: Learning weakly-supervised scene graph
  for complex visual grounding}.
\newblock \bibinfo{journal}{\emph{Neurocomputing}}  \bibinfo{volume}{413}
  (\bibinfo{year}{2020}), \bibinfo{pages}{51--60}.
\newblock


\bibitem[Ma et~al\mbox{.}(2021)]%
        {ma2021implicit}
\bibfield{author}{\bibinfo{person}{Lufan Ma}, \bibinfo{person}{Tiancai Wang},
  \bibinfo{person}{Bin Dong}, \bibinfo{person}{Jiangpeng Yan},
  \bibinfo{person}{Xiu Li}, {and} \bibinfo{person}{Xiangyu Zhang}.}
  \bibinfo{year}{2021}\natexlab{}.
\newblock \showarticletitle{Implicit feature refinement for instance
  segmentation}. In \bibinfo{booktitle}{\emph{Proceedings of the 29th ACM
  International Conference on Multimedia}}. \bibinfo{pages}{3088--3096}.
\newblock


\bibitem[Ma et~al\mbox{.}(2022)]%
        {ma2022open}
\bibfield{author}{\bibinfo{person}{Zongyang Ma}, \bibinfo{person}{Guan Luo},
  \bibinfo{person}{Jin Gao}, \bibinfo{person}{Liang Li}, \bibinfo{person}{Yuxin
  Chen}, \bibinfo{person}{Shaoru Wang}, \bibinfo{person}{Congxuan Zhang}, {and}
  \bibinfo{person}{Weiming Hu}.} \bibinfo{year}{2022}\natexlab{}.
\newblock \showarticletitle{Open-Vocabulary One-Stage Detection with
  Hierarchical Visual-Language Knowledge Distillation}. In
  \bibinfo{booktitle}{\emph{Proceedings of the IEEE/CVF Conference on Computer
  Vision and Pattern Recognition}}. \bibinfo{pages}{14074--14083}.
\newblock


\bibitem[Paszke et~al\mbox{.}(2019)]%
        {paszke2019pytorch}
\bibfield{author}{\bibinfo{person}{Adam Paszke}, \bibinfo{person}{Sam Gross},
  \bibinfo{person}{Francisco Massa}, \bibinfo{person}{Adam Lerer},
  \bibinfo{person}{James Bradbury}, \bibinfo{person}{Gregory Chanan},
  \bibinfo{person}{Trevor Killeen}, \bibinfo{person}{Zeming Lin},
  \bibinfo{person}{Natalia Gimelshein}, \bibinfo{person}{Luca Antiga},
  {et~al\mbox{.}}} \bibinfo{year}{2019}\natexlab{}.
\newblock \showarticletitle{Pytorch: An imperative style, high-performance deep
  learning library}.
\newblock \bibinfo{journal}{\emph{Advances in neural information processing
  systems}}  \bibinfo{volume}{32} (\bibinfo{year}{2019}),
  \bibinfo{pages}{8026--8037}.
\newblock


\bibitem[Plummer et~al\mbox{.}(2015)]%
        {plummer2015flickr30k}
\bibfield{author}{\bibinfo{person}{Bryan~A Plummer}, \bibinfo{person}{Liwei
  Wang}, \bibinfo{person}{Chris~M Cervantes}, \bibinfo{person}{Juan~C Caicedo},
  \bibinfo{person}{Julia Hockenmaier}, {and} \bibinfo{person}{Svetlana
  Lazebnik}.} \bibinfo{year}{2015}\natexlab{}.
\newblock \showarticletitle{Flickr30k entities: Collecting region-to-phrase
  correspondences for richer image-to-sentence models}. In
  \bibinfo{booktitle}{\emph{Proceedings of the IEEE international conference on
  computer vision}}. \bibinfo{pages}{2641--2649}.
\newblock


\bibitem[Qiu et~al\mbox{.}(2020)]%
        {qiu2020language}
\bibfield{author}{\bibinfo{person}{Heqian Qiu}, \bibinfo{person}{Hongliang Li},
  \bibinfo{person}{Qingbo Wu}, \bibinfo{person}{Fanman Meng},
  \bibinfo{person}{Hengcan Shi}, \bibinfo{person}{Taijin Zhao}, {and}
  \bibinfo{person}{King~Ngi Ngan}.} \bibinfo{year}{2020}\natexlab{}.
\newblock \showarticletitle{Language-Aware Fine-Grained Object Representation
  for Referring Expression Comprehension}. In
  \bibinfo{booktitle}{\emph{Proceedings of the 28th ACM International
  Conference on Multimedia}}. \bibinfo{pages}{4171--4180}.
\newblock


\bibitem[Radford et~al\mbox{.}(2021)]%
        {radford2021learning}
\bibfield{author}{\bibinfo{person}{Alec Radford}, \bibinfo{person}{Jong~Wook
  Kim}, \bibinfo{person}{Chris Hallacy}, \bibinfo{person}{Aditya Ramesh},
  \bibinfo{person}{Gabriel Goh}, \bibinfo{person}{Sandhini Agarwal},
  \bibinfo{person}{Girish Sastry}, \bibinfo{person}{Amanda Askell},
  \bibinfo{person}{Pamela Mishkin}, \bibinfo{person}{Jack Clark},
  {et~al\mbox{.}}} \bibinfo{year}{2021}\natexlab{}.
\newblock \showarticletitle{Learning transferable visual models from natural
  language supervision}.
\newblock \bibinfo{journal}{\emph{arXiv preprint arXiv:2103.00020}}
  (\bibinfo{year}{2021}).
\newblock


\bibitem[Rasheed et~al\mbox{.}(2022)]%
        {rasheed2022bridging}
\bibfield{author}{\bibinfo{person}{Hanoona Rasheed}, \bibinfo{person}{Muhammad
  Maaz}, \bibinfo{person}{Muhammad~Uzair Khattak}, \bibinfo{person}{Salman
  Khan}, {and} \bibinfo{person}{Fahad~Shahbaz Khan}.}
  \bibinfo{year}{2022}\natexlab{}.
\newblock \showarticletitle{Bridging the gap between object and image-level
  representations for open-vocabulary detection}.
\newblock \bibinfo{journal}{\emph{Conference on Neural Information Processing
  Systems}} (\bibinfo{year}{2022}).
\newblock


\bibitem[Ren et~al\mbox{.}(2017)]%
        {ren2015faster}
\bibfield{author}{\bibinfo{person}{Shaoqing Ren}, \bibinfo{person}{Kaiming He},
  \bibinfo{person}{Ross Girshick}, {and} \bibinfo{person}{Jian Sun}.}
  \bibinfo{year}{2017}\natexlab{}.
\newblock \showarticletitle{Faster R-CNN: Towards Real-Time Object Detection
  with Region Proposal Networks}.
\newblock \bibinfo{journal}{\emph{IEEE Transactions on Pattern Analysis \&
  Machine Intelligence}} \bibinfo{volume}{39}, \bibinfo{number}{06}
  (\bibinfo{year}{2017}), \bibinfo{pages}{1137--1149}.
\newblock


\bibitem[Schuster et~al\mbox{.}(2015)]%
        {schuster2015generating}
\bibfield{author}{\bibinfo{person}{Sebastian Schuster}, \bibinfo{person}{Ranjay
  Krishna}, \bibinfo{person}{Angel Chang}, \bibinfo{person}{Li Fei-Fei}, {and}
  \bibinfo{person}{Christopher~D Manning}.} \bibinfo{year}{2015}\natexlab{}.
\newblock \showarticletitle{Generating semantically precise scene graphs from
  textual descriptions for improved image retrieval}. In
  \bibinfo{booktitle}{\emph{Proceedings of the Fourth Workshop on Vision and
  Language}}. Citeseer, \bibinfo{pages}{70--80}.
\newblock


\bibitem[Shi et~al\mbox{.}(2018)]%
        {shi2018boosting}
\bibfield{author}{\bibinfo{person}{Hengcan Shi}, \bibinfo{person}{Hongliang
  Li}, \bibinfo{person}{Qingbo Wu}, \bibinfo{person}{Fanman Meng}, {and}
  \bibinfo{person}{King~N Ngan}.} \bibinfo{year}{2018}\natexlab{}.
\newblock \showarticletitle{Boosting Scene Parsing Performance via Reliable
  Scale Prediction}. In \bibinfo{booktitle}{\emph{2018 ACM Multimedia
  Conference on Multimedia Conference}}. ACM, \bibinfo{pages}{492--500}.
\newblock


\bibitem[Shi et~al\mbox{.}(2021)]%
        {shi2021simple}
\bibfield{author}{\bibinfo{person}{Jing Shi}, \bibinfo{person}{Yiwu Zhong},
  \bibinfo{person}{Ning Xu}, \bibinfo{person}{Yin Li}, {and}
  \bibinfo{person}{Chenliang Xu}.} \bibinfo{year}{2021}\natexlab{}.
\newblock \showarticletitle{A simple baseline for weakly-supervised scene graph
  generation}. In \bibinfo{booktitle}{\emph{Proceedings of the IEEE/CVF
  International Conference on Computer Vision}}. \bibinfo{pages}{16393--16402}.
\newblock


\bibitem[Shit et~al\mbox{.}(2022)]%
        {shit2022relationformer}
\bibfield{author}{\bibinfo{person}{Suprosanna Shit}, \bibinfo{person}{Rajat
  Koner}, \bibinfo{person}{Bastian Wittmann}, \bibinfo{person}{Johannes
  Paetzold}, \bibinfo{person}{Ivan Ezhov}, \bibinfo{person}{Hongwei Li},
  \bibinfo{person}{Jiazhen Pan}, \bibinfo{person}{Sahand Sharifzadeh},
  \bibinfo{person}{Georgios Kaissis}, \bibinfo{person}{Volker Tresp},
  {et~al\mbox{.}}} \bibinfo{year}{2022}\natexlab{}.
\newblock \showarticletitle{Relationformer: A unified framework for
  image-to-graph generation}. In \bibinfo{booktitle}{\emph{Computer
  Vision--ECCV 2022: 17th European Conference, Tel Aviv, Israel, October
  23--27, 2022, Proceedings, Part XXXVII}}. \bibinfo{pages}{422--439}.
\newblock


\bibitem[Tang et~al\mbox{.}(2021)]%
        {tang2021clip4caption}
\bibfield{author}{\bibinfo{person}{Mingkang Tang}, \bibinfo{person}{Zhanyu
  Wang}, \bibinfo{person}{Zhenhua Liu}, \bibinfo{person}{Fengyun Rao},
  \bibinfo{person}{Dian Li}, {and} \bibinfo{person}{Xiu Li}.}
  \bibinfo{year}{2021}\natexlab{}.
\newblock \showarticletitle{Clip4caption: Clip for video caption}. In
  \bibinfo{booktitle}{\emph{Proceedings of the 29th ACM International
  Conference on Multimedia}}. \bibinfo{pages}{4858--4862}.
\newblock


\bibitem[Vaswani et~al\mbox{.}(2017)]%
        {vaswani2017attention}
\bibfield{author}{\bibinfo{person}{Ashish Vaswani}, \bibinfo{person}{Noam
  Shazeer}, \bibinfo{person}{Niki Parmar}, \bibinfo{person}{Jakob Uszkoreit},
  \bibinfo{person}{Llion Jones}, \bibinfo{person}{Aidan~N Gomez},
  \bibinfo{person}{{\L}ukasz Kaiser}, {and} \bibinfo{person}{Illia
  Polosukhin}.} \bibinfo{year}{2017}\natexlab{}.
\newblock \showarticletitle{Attention is all you need}.
\newblock \bibinfo{journal}{\emph{Advances in neural information processing
  systems}}  \bibinfo{volume}{30} (\bibinfo{year}{2017}).
\newblock


\bibitem[Wang et~al\mbox{.}(2022)]%
        {wang2022point}
\bibfield{author}{\bibinfo{person}{Zheng Wang}, \bibinfo{person}{Zhenwei Gao},
  \bibinfo{person}{Xing Xu}, \bibinfo{person}{Yadan Luo}, \bibinfo{person}{Yang
  Yang}, {and} \bibinfo{person}{Heng~Tao Shen}.}
  \bibinfo{year}{2022}\natexlab{}.
\newblock \showarticletitle{Point to Rectangle Matching for Image Text
  Retrieval}. In \bibinfo{booktitle}{\emph{Proceedings of the 30th ACM
  International Conference on Multimedia}}. \bibinfo{pages}{4977--4986}.
\newblock


\bibitem[Xu et~al\mbox{.}(2017)]%
        {xu2017scene}
\bibfield{author}{\bibinfo{person}{Danfei Xu}, \bibinfo{person}{Yuke Zhu},
  \bibinfo{person}{Christopher~B Choy}, {and} \bibinfo{person}{Li Fei-Fei}.}
  \bibinfo{year}{2017}\natexlab{}.
\newblock \showarticletitle{Scene graph generation by iterative message
  passing}. In \bibinfo{booktitle}{\emph{Proceedings of the IEEE conference on
  computer vision and pattern recognition}}. \bibinfo{pages}{5410--5419}.
\newblock


\bibitem[Yang et~al\mbox{.}(2020)]%
        {yang2020graph}
\bibfield{author}{\bibinfo{person}{Sibei Yang}, \bibinfo{person}{Guanbin Li},
  {and} \bibinfo{person}{Yizhou Yu}.} \bibinfo{year}{2020}\natexlab{}.
\newblock \showarticletitle{Graph-structured referring expression reasoning in
  the wild}. In \bibinfo{booktitle}{\emph{Proceedings of the IEEE/CVF
  conference on computer vision and pattern recognition}}.
  \bibinfo{pages}{9952--9961}.
\newblock


\bibitem[Zang et~al\mbox{.}(2022)]%
        {zang2022open}
\bibfield{author}{\bibinfo{person}{Yuhang Zang}, \bibinfo{person}{Wei Li},
  \bibinfo{person}{Kaiyang Zhou}, \bibinfo{person}{Chen Huang}, {and}
  \bibinfo{person}{Chen~Change Loy}.} \bibinfo{year}{2022}\natexlab{}.
\newblock \showarticletitle{Open-Vocabulary DETR with Conditional Matching}.
\newblock \bibinfo{journal}{\emph{European Conference on Computer Vision}}
  (\bibinfo{year}{2022}).
\newblock


\bibitem[Zareian et~al\mbox{.}(2021)]%
        {zareian2021open}
\bibfield{author}{\bibinfo{person}{Alireza Zareian},
  \bibinfo{person}{Kevin~Dela Rosa}, \bibinfo{person}{Derek~Hao Hu}, {and}
  \bibinfo{person}{Shih-Fu Chang}.} \bibinfo{year}{2021}\natexlab{}.
\newblock \showarticletitle{Open-vocabulary object detection using captions}.
  In \bibinfo{booktitle}{\emph{Proceedings of the IEEE/CVF Conference on
  Computer Vision and Pattern Recognition}}. \bibinfo{pages}{14393--14402}.
\newblock


\bibitem[Zhang et~al\mbox{.}(2022)]%
        {zhang2022glipv2}
\bibfield{author}{\bibinfo{person}{Haotian Zhang}, \bibinfo{person}{Pengchuan
  Zhang}, \bibinfo{person}{Xiaowei Hu}, \bibinfo{person}{Yen-Chun Chen},
  \bibinfo{person}{Liunian~Harold Li}, \bibinfo{person}{Xiyang Dai},
  \bibinfo{person}{Lijuan Wang}, \bibinfo{person}{Lu Yuan},
  \bibinfo{person}{Jenq-Neng Hwang}, {and} \bibinfo{person}{Jianfeng Gao}.}
  \bibinfo{year}{2022}\natexlab{}.
\newblock \showarticletitle{Glipv2: Unifying localization and vision-language
  understanding}.
\newblock \bibinfo{journal}{\emph{Conference on Neural Information Processing
  Systems}} (\bibinfo{year}{2022}).
\newblock


\bibitem[Zhao et~al\mbox{.}(2022)]%
        {zhao2022exploiting}
\bibfield{author}{\bibinfo{person}{Shiyu Zhao}, \bibinfo{person}{Zhixing
  Zhang}, \bibinfo{person}{Samuel Schulter}, \bibinfo{person}{Long Zhao},
  \bibinfo{person}{BG Vijay~Kumar}, \bibinfo{person}{Anastasis Stathopoulos},
  \bibinfo{person}{Manmohan Chandraker}, {and} \bibinfo{person}{Dimitris~N
  Metaxas}.} \bibinfo{year}{2022}\natexlab{}.
\newblock \showarticletitle{Exploiting unlabeled data with vision and language
  models for object detection}. In \bibinfo{booktitle}{\emph{European
  Conference on Computer Vision}}. \bibinfo{pages}{159--175}.
\newblock


\bibitem[Zhong et~al\mbox{.}(2021)]%
        {zhong2021learning}
\bibfield{author}{\bibinfo{person}{Yiwu Zhong}, \bibinfo{person}{Jing Shi},
  \bibinfo{person}{Jianwei Yang}, \bibinfo{person}{Chenliang Xu}, {and}
  \bibinfo{person}{Yin Li}.} \bibinfo{year}{2021}\natexlab{}.
\newblock \showarticletitle{Learning to generate scene graph from natural
  language supervision}. In \bibinfo{booktitle}{\emph{Proceedings of the
  IEEE/CVF International Conference on Computer Vision}}.
  \bibinfo{pages}{1823--1834}.
\newblock


\bibitem[Zhong et~al\mbox{.}(2022)]%
        {zhong2022regionclip}
\bibfield{author}{\bibinfo{person}{Yiwu Zhong}, \bibinfo{person}{Jianwei Yang},
  \bibinfo{person}{Pengchuan Zhang}, \bibinfo{person}{Chunyuan Li},
  \bibinfo{person}{Noel Codella}, \bibinfo{person}{Liunian~Harold Li},
  \bibinfo{person}{Luowei Zhou}, \bibinfo{person}{Xiyang Dai},
  \bibinfo{person}{Lu Yuan}, \bibinfo{person}{Yin Li}, {et~al\mbox{.}}}
  \bibinfo{year}{2022}\natexlab{}.
\newblock \showarticletitle{Regionclip: Region-based language-image
  pretraining}. In \bibinfo{booktitle}{\emph{Proceedings of the IEEE/CVF
  Conference on Computer Vision and Pattern Recognition}}.
  \bibinfo{pages}{16793--16803}.
\newblock


\bibitem[Zhou et~al\mbox{.}(2022)]%
        {zhou2022detecting}
\bibfield{author}{\bibinfo{person}{Xingyi Zhou}, \bibinfo{person}{Rohit
  Girdhar}, \bibinfo{person}{Armand Joulin}, \bibinfo{person}{Phillip
  Kr{\"a}henb{\"u}hl}, {and} \bibinfo{person}{Ishan Misra}.}
  \bibinfo{year}{2022}\natexlab{}.
\newblock \showarticletitle{Detecting twenty-thousand classes using image-level
  supervision}.
\newblock \bibinfo{journal}{\emph{European Conference on Computer Vision}}
  (\bibinfo{year}{2022}).
\newblock


\bibitem[Zhu et~al\mbox{.}(2021)]%
        {zhu2021deformable}
\bibfield{author}{\bibinfo{person}{Xizhou Zhu}, \bibinfo{person}{Weijie Su},
  \bibinfo{person}{Lewei Lu}, \bibinfo{person}{Bin Li},
  \bibinfo{person}{Xiaogang Wang}, {and} \bibinfo{person}{Jifeng Dai}.}
  \bibinfo{year}{2021}\natexlab{}.
\newblock \showarticletitle{Deformable DETR: Deformable Transformers for
  End-to-End Object Detection}. In \bibinfo{booktitle}{\emph{International
  Conference on Learning Representations}}.
\newblock


\end{thebibliography}

	\end{document}